\documentclass[preprint, 3p, review,12pt]{elsarticle}
\usepackage[numbers]{natbib}
\usepackage{graphicx}

\usepackage{amssymb}
\usepackage{amsmath}
\usepackage{needspace}
\usepackage{adjustbox}  % To resize tables
\usepackage{placeins}
\usepackage{booktabs} % For \toprule, \midrule, and \bottomrule
\usepackage{tabularx, booktabs}
\usepackage{graphicx} % in the preamble
\usepackage{algpseudocode}
\usepackage{hyperref}    % For clickable references
\usepackage[ruled,vlined]{algorithm2e}
\usepackage{fancyhdr}  % Load the fancyhdr package
\SetAlgoVlined  % Ensures long algorithms break properly
\SetAlgoNlRelativeSize{-1}  % Adjusts line number size to avoid layout issues
\DontPrintSemicolon  % Can sometimes help prevent unnecessary spacing issues

\journal{Discover Artificial Intelligence}

\begin{document}

%\begin{frontmatter}

%% Title, authors and addresses

%% use the tnoteref command within \title for footnotes;
%% use the tnotetext command for theassociated footnote;
%% use the fnref command within \author or \affiliation for footnotes;
%% use the fntext command for theassociated footnote;
%% use the corref command within \author for corresponding author footnotes;
%% use the cortext command for theassociated footnote;
%% use the ead command for the email address,
%% and the form \ead[url] for the home page:
%% \title{Title\tnoteref{label1}}
%% \tnotetext[label1]{}
%% \author{Name\corref{cor1}\fnref{label2}}
%% \ead{email address}
%% \ead[url]{home page}
%% \fntext[label2]{}
%% \cortext[cor1]{}
%% \affiliation{organization={},
%%             addressline={},
%%             city={},
%%             postcode={},
%%             state={},
%%             country={}}
%% \fntext[label3]{}

\title{A Structured Review of Underwater Object Detection Challenges and Solutions: From Traditional to Large Vision Language Models}

%% use optional labels to link authors explicitly to addresses:
%% \author[label1,label2]{}
%% \affiliation[label1]{organization={},
%%             addressline={},
%%             city={},
%%             postcode={},
%%             state={},
%%             country={}}
%%
%% \affiliation[label2]{organization={},
%%             addressline={},
%%             city={},
%%             postcode={},
%%             state={},
%%             country={}}

%% Authors:
\author[inst1]{Edwine Nabahirwa}
\author[inst1]{Wei Song \corref{cor1}}
\author[inst1]{Minghua Zhang}
\author[inst1]{Yi Fang}
\author[inst1]{Zhou Ni}

%% Single affiliation:
\affiliation[inst1]{organization={Shanghai Ocean University},%Department and Organization
            city={Shanghai},
            postcode={201306},
            country={China}}
%% Corresponding author footnote:
\cortext[cor1]{Corresponding author: wsong@shou.edu.cn}

%% Abstract

\begin{abstract}
%% Text of abstract
Underwater object detection (UOD) is vital to diverse marine applications, including oceanographic research, underwater robotics, and marine conservation. However, UOD faces numerous challenges that compromise its performance. Over the years, various methods have been proposed to address these issues, but they often fail to fully capture the complexities of underwater environments. This review systematically categorizes UOD challenges into five key areas: Image quality degradation, target-related issues, data-related challenges, computational and processing constraints, and limitations in detection methodologies.
To address these challenges, we analyze the progression from traditional image processing and object detection techniques to modern approaches. Additionally, we explore the potential of large vision-language models (LVLMs) in UOD, leveraging their multi-modal capabilities demonstrated in other domains. We also present case studies, including synthetic dataset generation using DALL-E 3 and fine-tuning Florence-2 LVLM for UOD. This review identifies three key insights: (i) Current UOD methods are insufficient to fully address challenges like image degradation and small object detection in dynamic underwater environments. (ii) Synthetic data generation using LVLMs shows potential for augmenting datasets but requires further refinement to ensure realism and applicability. (iii) LVLMs hold significant promise for UOD, but their real-time application remains under-explored, requiring further research on optimization techniques.
\end{abstract}

%% Keywords

%\end{frontmatter}

%% Add \usepackage{lineno} before \begin{document} and uncomment 
%% following line to enable line numbers
%% \linenumbers

%% main text
%%
\maketitle

\section{INTRODUCTION}
The ocean, which covers approximately 70\% of the Earth's surface, hosts 97\% of its water, and sustains vital ecosystems.
%The ocean, which covers approximately 70\% of the Earth's surface and hosts 97\% of its water, is one of the most crucial, yet least understood environments on the planet. It is a major source of global oxygen and harbors essential ecosystems, which regulate climate and support marine food webs. 
Despite its significance, the underwater world remains largely overlooked as a result of the challenging conditions that hinder traditional research methods. Historically, the study of marine ecosystems relied on labor intensive research \cite{mallet_underwater_2014}, which provided limited data and had a high error margin. In recent years, advances in autonomous and remotely operated vehicles (AUVs and ROVs) have revolutionized underwater exploration. These technologies, equipped with object detection systems, now allow real-time monitoring, which includes capturing images of marine organisms, environmental conditions, and even assessing biodiversity \cite{carlucho_adaptive_2018}, \cite{sahoo_advancements_2019}. However, the quality of images and videos captured underwater remains a significant obstacle. Light absorption, scattering, and water-related distortions, such as haze and color shifts \cite{chiang_underwater_2012}, create noisy low-contrast images, further compounded by complex underwater backgrounds and camera motion. These challenges call for advanced detection techniques capable of accurately identifying and localizing objects despite underwater noise.

Efficient underwater object detection (UOD) is crucial for a variety of marine applications, including biodiversity monitoring, conservation efforts, and resource management. Object detection facilitates identification and counting of marine species, supporting crucial research on ecosystem health and enabling sustainable management. However, achieving robust performance in underwater environments is complex due to several factors, including the lack of sufficient annotated underwater data, which limits the ability of machine learning models to generalize \cite{er_research_2023}. Deep learning-based models require large and diverse datasets to function effectively, but the scarcity of such datasets in underwater research often results in overfitting and poor generalization.
More so, state-of-the-art detection frameworks often face significant challenges due to the inherent difficulty of detecting small underwater objects or those that blend seamlessly into the environment. These challenges are further compounded by label noise caused by poor image quality and imbalanced datasets, where certain object classes are heavily overrepresented while others remain underrepresented. These factors pose a significant challenge for real-world applications of UOD. As the demand for precise, real-time monitoring of marine environments grows, overcoming these challenges becomes increasingly urgent. In this context, the recent advent of LVLMs \cite{zhang_vision-language_2023} has sparked excitement in the research community. A couple of LVLMs have been developed with capability in tasks such as captioning, VQA, and object localization, which are essential capabilities for UOD \cite{caffagni_revolution_2024}. Their ability to learn from vast amounts of multimodal data makes them an ideal candidate for addressing the inherent complexities in underwater environments.

Several reviews have been conducted on UOD, focusing on various aspects ranging from deep learning techniques to dataset analysis and unique challenges in underwater environments. Research has focused on the transition from conventional methods to deep learning for real-time and accurate detection and future trends \cite{blanc-talon_deep_2017}, \cite{wang_review_2022}, \cite{fayaz_underwater_2022}, \cite{jian_underwater_2024}, others have investigated AI-driven strategies, dataset evaluation, and error diagnosis tools \cite{chen_underwater_2024}, as well as broader applications of detection algorithms in wireless sensor networks \cite{khan_underwater_2024}, collectively providing a holistic understanding of the field. Other reviews have analysed image enhancement methods for improving object detection \cite{xu_systematic_2023}, introduced the RUOD dataset to address light interference and color casts using joint learning \cite{fu_rethinking_2023}, highlighted issues like image degradation, small object detection, and real-time performance in underwater settings \cite{er_research_2023} and emphasized the role of AUVs in dynamic UOD\cite{gomes_robust_2020}.

\begin{figure}[h]
  \centering
  \includegraphics[width=0.9\linewidth]{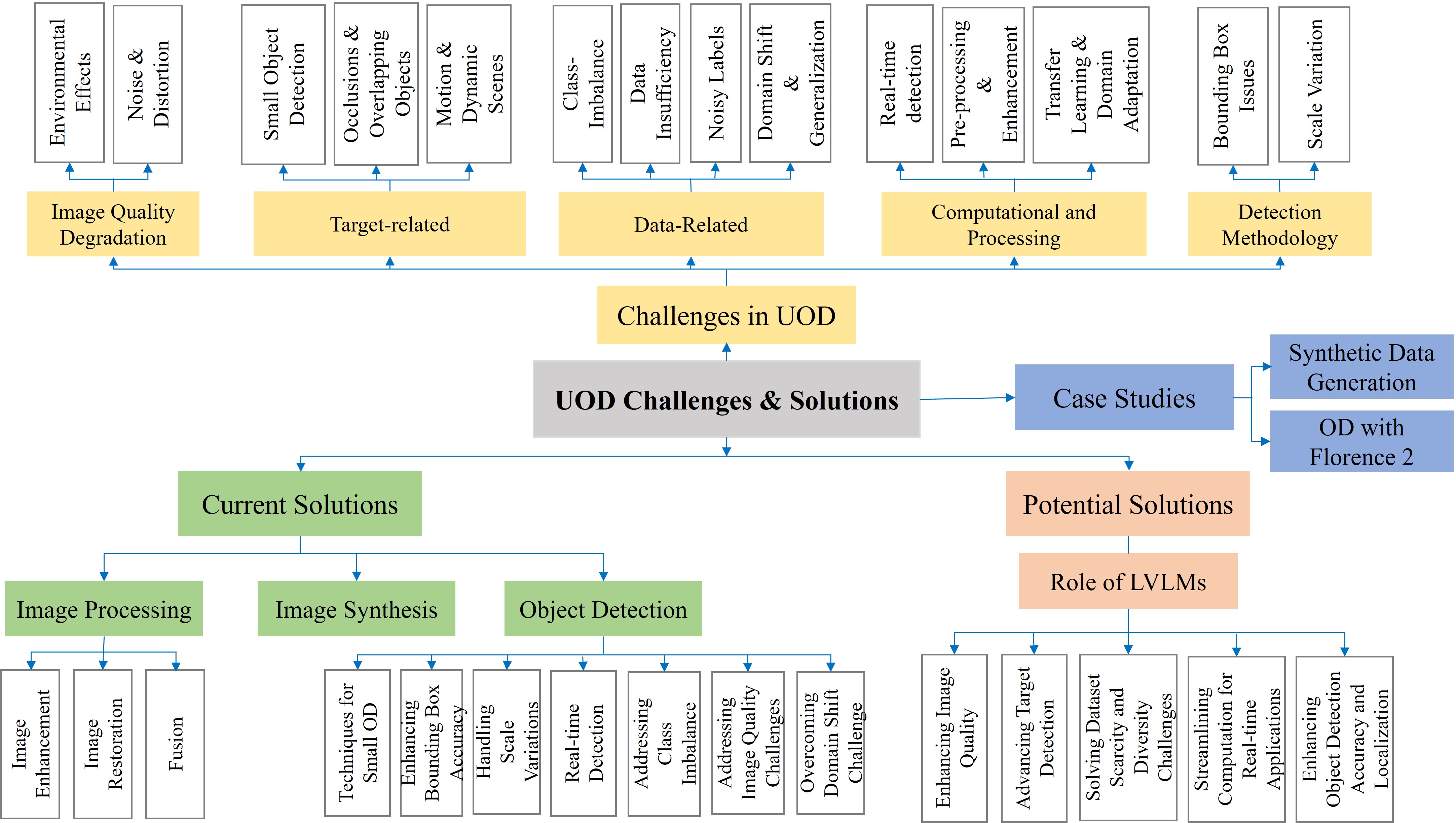}
  \caption{Taxonomy of UOD Challenges and Solutions}
  \label{fig:taxonomy} % Added label
\end{figure}

Several issues have not been thoroughly addressed in existing reviews on UOD: 1) Most reviews focus primarily on traditional techniques, leaving a gap in exploring the integration of modern advancements, such as LVLMs, in UOD; 2) While challenges like image quality degradation, target-related issues, and computational constraints are discussed, few reviews offer a comprehensive exploration of both the challenges and innovative solutions that could address them. As the primary contribution of our review, we aim to bridge these gaps by presenting a holistic and innovative perspective on UOD. The major contributions of this paper are threefold:

1)	This review provides a detailed and structured analysis of the multifaceted challenges in UOD, systematically categorizing them and exploring both traditional and modern solutions. It offers a comprehensive understanding of the complexities inherent in UOD and highlights emerging trends and future research directions, as taxonomized in Fig.~\ref{fig:taxonomy}.

2)	A significant contribution of this review is its focus on the potential of LVLMs in addressing the unique challenges of UOD. It showcases the evolution and breakthroughs of LVLMs, their success in other vision tasks such as VQA and image captioning, and explores how they can be fine-tuned or adapted for UOD applications.

3)	The review introduces case studies on underwater synthetic data generation using DALL-E 3 and the efficient fine-tuning of Florence-2 LVLM with LoRA for object detection. These case studies demonstrate how these novel techniques can address critical data limitations, such as insufficient training data, class imbalance, and domain shift, offering scalable solutions for developing robust UOD models.

This paper is organized as follows: Section 2 categorizes the existing challenges in UOD, while Section 3 explores progress in mitigating these issues. In Section 4, we delve into the role of LVLMs, focusing on breakthroughs and their application to UOD. Section 5 presents case studies on synthetic data generation and LVLM fine-tuning for UOD. Finally, Section 6 outlines future research directions and concludes the study.

\section{CHALLENGES IN UNDERWATER OBJECT DETECTION}
UOD is a complex task that involves a range of challenges unique to the submerged environment. This section delves into these key challenges, providing a foundation for exploring potential solutions. The challenges are grouped into meaningful categories, with specific issues classified under each. More so, as illustrated in Table.~\ref{tab:categ}., and  Fig.~\ref{fig:duo_challenges}, several challenges faced by UOD are highlighted using examples from the DUO public dataset~\cite{liu_dataset_2021}:

\begin{figure}[h]
  \centering
  \includegraphics[width=0.8\linewidth]{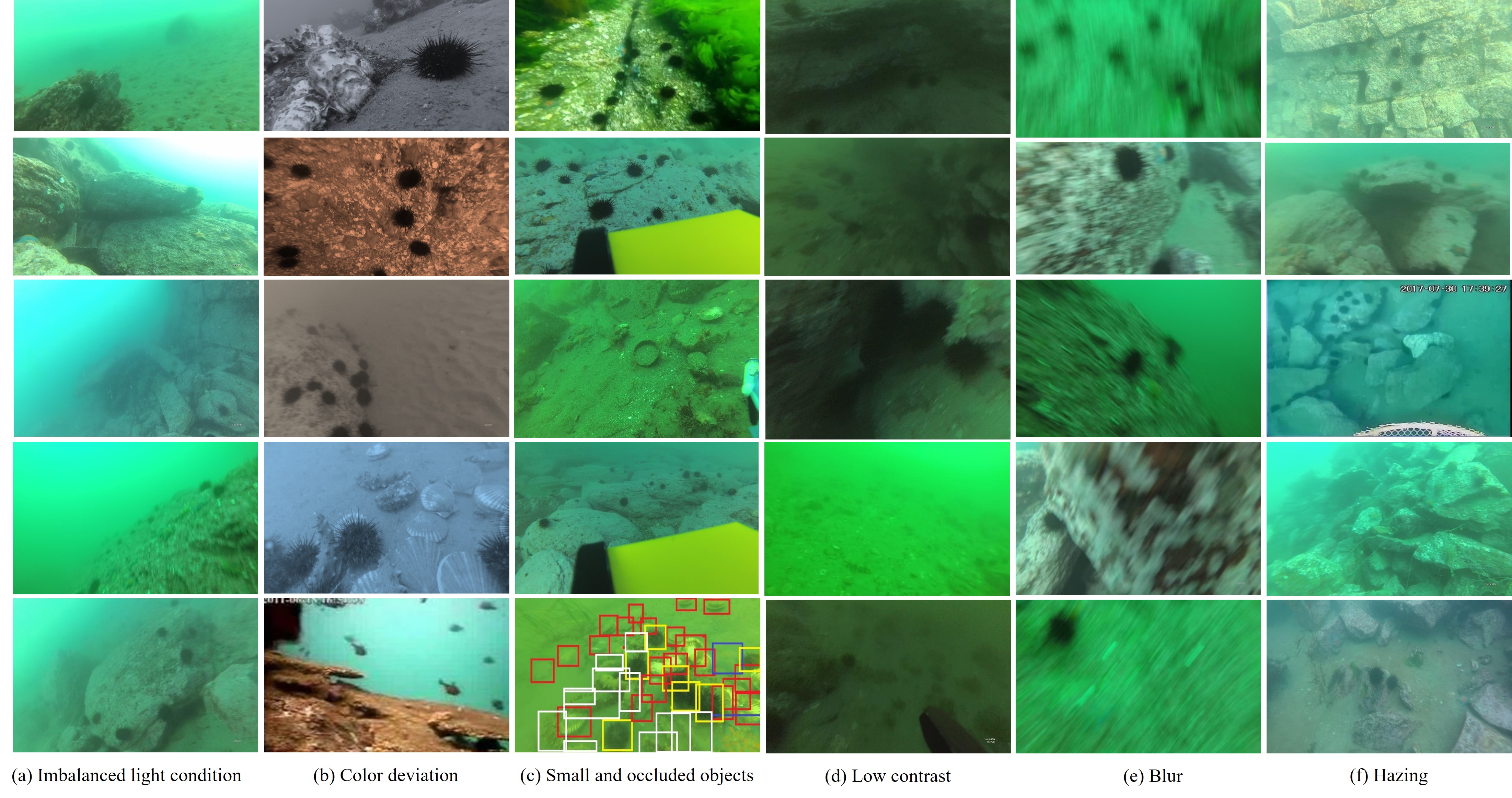}
  \caption{Some of the Challenges Faced by UOD from DUO Dataset \cite {liu_dataset_2021} }
  \label{fig:duo_challenges}
\end{figure}
\FloatBarrier % Ensures no floats move beyond this point

\subsection{Image Quality Degradation Challenges}
The degradation of image quality is caused by both environmental factors and external interference. These challenges affect the clarity, contrast, and usability of underwater imagery, thereby hindering accurate object detection and classification:

\begin{itemize}
\item\textbf{Environmental Effects:}
The inherent properties of underwater environments, such as light scattering, absorption, and refraction, significantly degrade image quality. Poor light conditions and complete darkness at greater depths lead to low contrast, blurry textures, and color distortions. Turbidity and haze-like effects further obscure visibility, while the scattering phenomenon introduces imbalanced lighting across the scene. These environmental factors create an overall reduction in image resolution and visual fidelity, complicating the detection of objects against the complex underwater background \cite{er_research_2023}, \cite{cherian_deep_2021}, \cite{joshi_underwater_2024}.

\item\textbf{Noise and Distortion:}
External interference, such as noise from shipping activity, wave disturbances, and ambient vibrations, adds another layer of complexity to image degradation. This noise often manifests as blurred or distorted images with random artifacts, which can vary depending on the frequency spectrum of the noise source \cite{yuan_survey_2022}, \cite{adeoluwa_evaluation_2023}, \cite{yang_underwater_2023}. 

\end{itemize}

\subsection{Target-Related Challenges}
These challenges arise from the intrinsic characteristics of objects and their interactions with the underwater environment, which directly impacts the ability of detection systems to accurately locate and identify targets, particularly in complex and dynamic underwater scenes:
\begin{itemize}
\item\textbf{Small Object Detection:}
Detecting small or tiny objects in underwater environments is difficult due to their minimal size and the overwhelming presence of surrounding elements. Scale variations further complicate detection, as objects of interest may range from microscopic marine organisms to larger entities like fish or debris. 
Additionally, inter-class similarity between objects (e.g., different species), or between objects and the background, often caused by camouflage or similar appearances, reduces the distinguishability of targets, making accurate detection a formidable task \cite{qi_underwater_2022}, \cite{zhang_vision-language_2023}.
%Additionally, inter-class similarity between objects (e.g., different species) and the background, often caused by camouflage or similar appearances, reduces the distinguishability of targets, making accurate detection a formidable task \cite{qi_underwater_2022}, \cite{zhang_vision-language_2023}.

\item\textbf{Occlusions and Overlapping Objects:}
Underwater scenes frequently feature partial or full occlusions, where objects of interest are blocked by marine life, debris, or other environmental elements. These occlusions reduce the visibility of targets and challenge detection models. Overlapping objects, particularly small ones, add another layer of complexity. Traditional bounding box approaches often struggle to differentiate and separate multiple objects within the same region, leading to inaccurate localization or identification \cite{wang_underwater_2023}.

\item\textbf{Motion and Dynamic Scenes:}
The dynamic nature of underwater environments introduces additional challenges. Moving targets, such as fish or vehicles, require detection systems with fast response times and robust motion-handling capabilities. Rapid changes in position, orientation, or visibility of targets can lead to incomplete or inaccurate detections if the system fails to adapt quickly \cite{han_mat_2022}, \cite{lee_artificial_2023}.
\end{itemize}

\subsection{Data-Related Challenges}
These challenges arise from the inherent limitations in the availability, quality, and consistency of training datasets, which normally affect model performance, generalization, and robustness:

\begin{itemize}
\item\textbf{Dataset Insufficiency:}
The lack of sufficient, quality and diverse underwater image datasets remains a critical bottleneck for training and evaluating detection models. Existing datasets often contain a limited number of samples \cite{yu_treat_2023}, which fails to represent the wide variety of underwater conditions, environments, and target types encountered in real-world applications.

\item\textbf{Class Imbalance:}
This happens when certain categories of underwater objects (e.g., common species) being overrepresented, while others (e.g., rare species) are severely underrepresented. This imbalance can bias detection models, leading to poor performance on underrepresented classes \cite{yu_treat_2023}, \cite{chen_underwater_2024}.

\item\textbf{Noisy Labels:}
Noisy or inaccurate annotations in training datasets introduce significant challenges for supervised learning \cite{saoud_adod_2023}. Errors in labeling, such as incorrect bounding boxes or misclassified objects, compromise the reliability of the training data and propagate inaccuracies into the detection model. This is particularly problematic in underwater settings, where complex scenes and low image quality often make manual annotation difficult and error-prone \cite{agrawal_syn2real_2024}.

\item\textbf{Domain Shift and Generalization:}
Underwater environments vary widely in terms of visibility, lighting, turbidity, and biological diversity across different locations and depths. These variations create a domain shift between the conditions represented in training datasets and those encountered during real-world deployment. Consequently, models trained on specific datasets often fail to generalize effectively to unseen environments, leading to significant drops in detection performance. Differences between training and testing domains, such as between clear and turbid waters or tropical and deep-sea environments, exacerbate this issue and limit the scalability of detection models \cite{agrawal_syn2real_2024},\cite{walker_underwater_2024}. Publicly available UOD datasets and their limitations are summarized in Table.~\ref{tab:dataset_sample}.

\end{itemize}

\begin{table}[h]
  \centering
  \caption{A Summary of Popular Annotated Underwater Image Datasets}
  \label{tab:dataset_sample}
  \begin{adjustbox}{max width=\textwidth}
    \begin{tabular}{>{\centering\arraybackslash}m{3cm} >{\centering\arraybackslash}m{3cm} m{9cm}}
      \toprule
      \textbf{Dataset} & \textbf{Image Number} & \textbf{Limitations} \\
      \midrule
      RUOD \cite{fu_rethinking_2023} & 14,000 & Limited number of object classes, potential class imbalance, and lack of diversity in underwater environments. \\
      RF100 \cite{ciaglia_roboflow_2022} & 7,600 & Limited diversity, potential class imbalance, inconsistent annotations, and small size. \\
      UDD \cite{liu_new_2022} & 2,227 & Lacks diversity, contains many repeated images, and overlaps with other public datasets. \\
      UODD \cite{jiang_underwater_2021} & 3,194 & Limited diversity of objects, class imbalance, low-quality images, incomplete annotations, and narrow scope of environments. \\
      DUO \cite{liu_dataset_2021} & 7,782 & Collected by re-annotating URPC and UDD datasets; lacks diversity, many repeated/common images. \\
      URPC (2017, 2018, 2019, 2020DL, 2020ZJ) \cite{chen_mmdetection_2019} & (18,640, 3,701, 5,786, 8,975, 7,543) & Lacks annotation files for testing; limited diversity. \\
      Brackish \cite{Pedersen_2019_CVPR_Workshops} & 14,518 & Narrow object class range, poor image quality, environmental bias specific to brackish water, class imbalance. \\
      WildFish \cite{zhuang_wildfish_2018} & 54,459 & Poor image quality and limited diversity. \\
      F4K-species \cite{international_association_for_pattern_recognition_21st_2012} & 27,370 & Lacks ready-made annotations, significant class imbalance, limited diversity. \\
      \bottomrule
    \end{tabular}
  \end{adjustbox}
\end{table}

\subsection{Computational and Processing Challenges}
These challenges are as a result of limitations of hardware platforms and the complexity of algorithms required to achieve high detection accuracy in real-time applications, which impacts the speed, efficiency, and practicality of deploying detection systems in resource-constrained underwater environments:
\begin{itemize}
    \item\textbf{Real-Time Detection:}
    This is crucial for applications like AUVs and marine robotics, where decisions need to be made instantly. The embedded platforms commonly used in underwater systems often have limited computing power and memory capacity. These constraints make it challenging to deploy complex detection models that require high computational resources. Accurate detection models, such as those based on deep learning, typically involve intensive computations, which may lead to delays or infeasibility in real-time scenarios \cite{zhao_improved_2022}.
    
    \item\textbf{Preprocessing and Enhancement:}
    Underwater images frequently require preprocessing and enhancement to improve overall quality before detection algorithms can effectively identify objects. This preprocessing step can be computationally expensive, especially when applied to high-resolution images or videos. Furthermore, while image enhancement can produce visually appealing results, it may inadvertently distort critical features, making images less suitable for object detection algorithms that rely on accurate and consistent representations \cite{xu_systematic_2023}, \cite{zhao_lightweight_2024}.
    
    \item\textbf{Transfer Learning and Domain Adaptation:}
    Given the domain-specific challenges of underwater environments, transfer learning and domain adaptation are often necessary to adapt pre-trained models to new underwater datasets. These techniques involve fine-tuning or retraining models on target datasets to bridge gaps between source and target domains. However, these processes are computationally intensive and require significant resources, particularly when dealing with large-scale models or diverse target environments \cite{menke_improving_2022}.
\end{itemize}

\begin{table}[h]
  \centering
  \caption{Categorization of UOD Challenges with Associated Issues}
  \label{tab:categ}
  \begin{adjustbox}{max width=\textwidth}
    \begin{tabular}{>{\raggedright\arraybackslash}m{4cm} m{11cm}}
      \toprule
      \textbf{Category} & \textbf{Specific Challenges} \\
      \midrule
      Image quality & Low light, scattering, blurring, texture loss, turbidity, color distortion, shipping, wave noise, frequency-specific filtering \\
      Target-related & Small objects, occlusions, overlapping objects, motion and dynamics \\
      Data-related & Insufficient datasets, class imbalance, noisy labels, domain shift, poor generalization \\
      Computational and Processing & Limited computing power, real-time detection requirements, preprocessing complexity \\
      Detection methodology & Overlap issues, bounding box limitations, scale variations \\
      \midrule
      \textbf{Challenges} & \textbf{Associated Issues} \\
      \midrule
      Low light & Poor visibility, color distortion, scattering \\
      Small object detection & Difficulty in identifying small or tiny objects, scale variations \\
      Noise & Blurring, distorted images, interference from thermal and ambient noise \\
      Data insufficiency & Lack of diverse, balanced, and high-quality datasets \\
      Real-time constraints & Limited computational resources on marine robots \\
      \bottomrule
    \end{tabular}
  \end{adjustbox}
\end{table}

\subsection{Detection Methodology Challenges}
Challenges under this category stem from inherent limitations in traditional and modern detection approaches, which struggle to adapt to the unique complexities of underwater environments hence affecting the accuracy and reliability of object detection.
\begin{itemize}
\item\textbf{Bounding Box Issues:}
Most detection methods rely heavily on bounding boxes to locate and classify objects within an image \cite{hao_research_2021}. However, these methods face significant difficulties in underwater environments where objects often overlap or occlude each other. Small, closely packed objects, such as schools of fish or clustered debris, are particularly challenging to differentiate, leading to misclassification or undercounting. In many cases, multiple overlapping objects are inaccurately identified as a single entity within the bounding box, reducing detection precision and limiting the system's ability to handle complex underwater scenes effectively \cite{ling_underwater_2023}.

\item\textbf{Scale Variation:}
Underwater objects can vary dramatically in size, ranging from tiny marine organisms to larger structures like shipwrecks or coral reefs. This significant variation in scale complicates detection, as a single model must effectively detect both large and small objects within the same scene. Many traditional and even modern detection methods struggle to maintain consistent accuracy across these scales, with smaller objects often being overlooked or misclassified due to insufficient resolution in feature extraction layers \cite{xu_scale-aware_2021}, \cite{guo_robust_2023}. 
\end{itemize}

\section{PROGRESS IN MITIGATING CHALLENGES IN UNDERWATER OBJECT DETECTION}

\begin{figure}[h]
  \centering
  \includegraphics[width=0.8\linewidth]{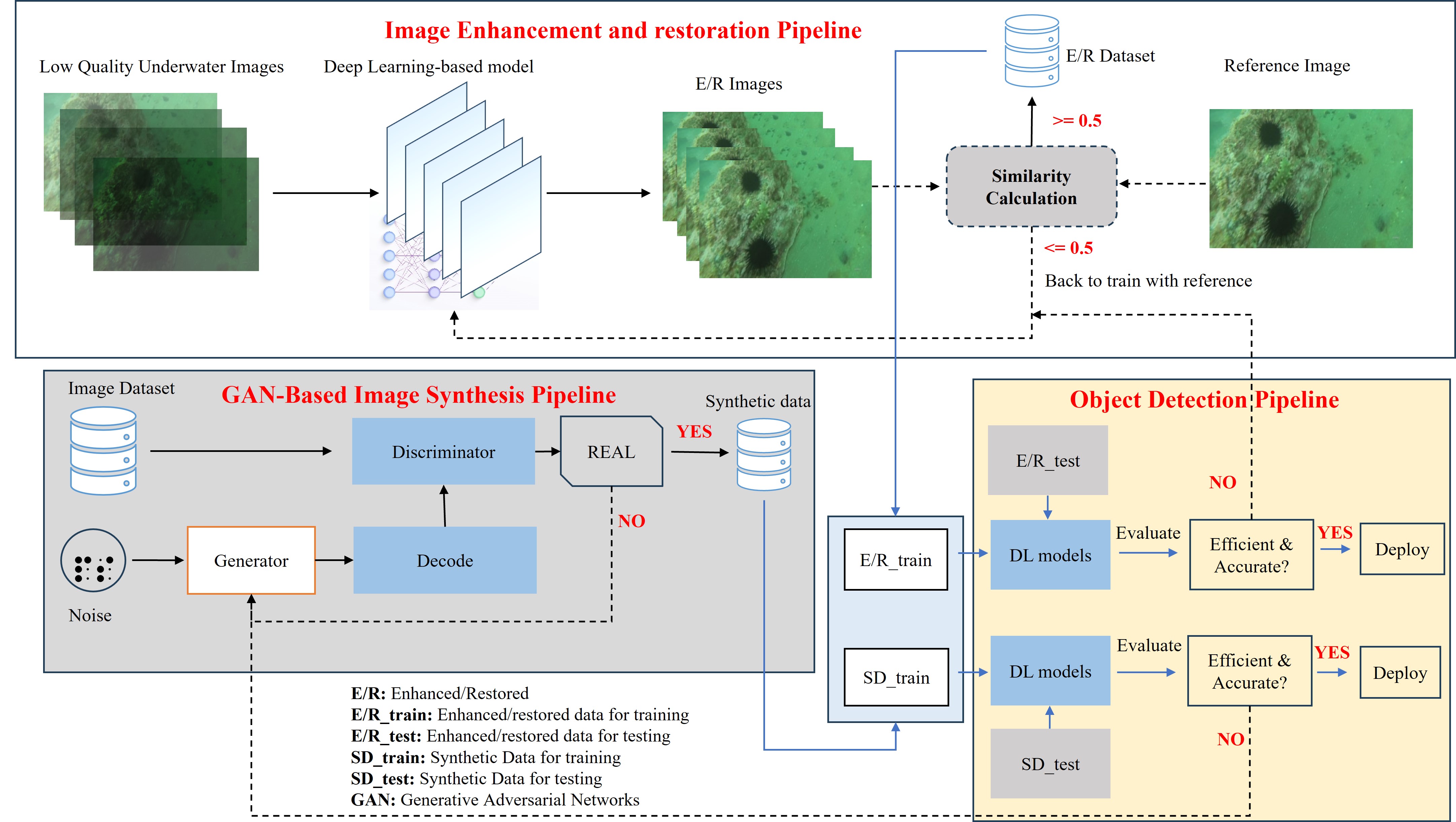}
  \caption{The relationship Between Image Enhancement, Restoration, Image Synthesis, and Robust Object Detection}
  \label{fig:enhancement and restoration}
\end{figure}

The field of UOD has seen significant advancements in addressing a range of complex challenges. Various model architectures and methodologies, from image processing, through image synthesis to detection techniques have been developed to tackle these issues. Fig.~\ref{fig:enhancement and restoration}, demonstrates the relationship between image enhancement, restoration, image synthesis, and robust object detection, highlighting how these processes interact to improve detection accuracy and efficiency through iterative feedback mechanisms.Image synthesis helps to generate data to train the detection model, while enhancement refines input image quality. The pipeline evaluates detection results, providing feedback to optimize synthesis parameters or enhancement settings. If results are suboptimal, adjustments are made iteratively, ensuring higher-quality inputs and more robust detection performance.
\subsection{Image Processing for Quality Underwater Object Detection}
This involves techniques to enhance and restore underwater images, addressing challenges like poor visibility, color degradation, and noise.
\subsubsection{Image Enhancement}
Methods here focus on improving visual quality, such as boosting contrast, brightness, and colors. Deep information can be found in a comprehensive review of underwater optical image enhancement \cite{shuang_algorithms_2024}.

\textbf{Super Resolution:}
A super-resolution method by Gao et al. \cite{gao_dae-gan_2024} use a synthetic dataset to enhance the BSRGAN architecture with an energy-based attention mechanism and a weighted fusion strategy integrating adversarial, reconstruction, and perceptual losses, showing significant gains in PSNR (+0.85 dB) and UIQM (+0.19). Also, an iterative framework \cite{singh_ida-uie_2024} uses condition-specific deep networks to detect and resolve image degradation, where a tailored enhancement network is applied based on the detected degradation, and degradation-specific datasets were constructed from UIEB and EUVP for training. 

\textbf{Color Calibration and Contrast Optimization:}
Underwater image enhancement addresses challenges like color distortion, low contrast, and poor visibility caused by light scattering and absorption. Various methods, including single-image approaches, color correction, and advanced fusion strategies, improve image clarity and utility for applications like object detection and marine exploration. Key techniques, their challenges, and improvements are reviewed in this section, as shown in Table.~\ref{tab:calib_contrast}.

\begin{table}[h]
  \centering
  \caption{Summary of Color Calibration and Contrast Optimization Techniques for Underwater Image Enhancement}
  \label{tab:calib_contrast}
  \begin{adjustbox}{max width=\textwidth}
  \renewcommand{\arraystretch}{1.2} % 行距放大到 1.5 倍
    \begin{tabular}{>{\raggedright\arraybackslash}m{0.25\textwidth} >{\raggedright\arraybackslash}m{0.52\textwidth} >{\raggedright\arraybackslash}m{0.23\textwidth}}
      \toprule
      \textbf{Model} & \textbf{Key Techniques/Modules} & \textbf{Addressed Challenges} \\
      \midrule
      Single-image enhancement with fusion \cite{ancuti_color_2018} & Color compensation, white balancing, edge-aware weight maps, multiscale fusion & Medium scattering, absorption, low contrast \\
      Color correction and adaptive contrast enhancement \cite{zhang_color_2021}, \cite{fan_innovative_2025} & Color channel compensation, adaptive contrast enhancement, unsharp masking & Color cast, low contrast \\
      Enhanced DCP with wavelet fusion \cite{zhu_underwater_2021} & Color compensation, white balance, improved dark channel prior (DCP), unsharp masking (USM), wavelet fusion & Color cast, low visibility, few edge details \\
      WWPF (Weighted Wavelet Visual Perception Fusion) \cite{zhang_underwater_2024} & Attenuation-map-guided color correction, global contrast enhancement, local contrast enhancement, wavelet fusion & Scattering, absorption, color distortion, low contrast \\
      ICSP-guided variational framework \cite{hou_non-uniform_2024} & Illumination channel sparsity prior (ICSP), Retinex theory-based variational model, ADMM algorithm & Insufficient light, non-uniform illumination \\
      UnitModule \cite{liu_unitmodule_2024} & Plug-and-play joint image enhancement, unsupervised learning loss, color cast predictor, UCRT (Underwater Color Random Transfer) & Underwater image degradation, color cast issues \\
      PDR with multi-priority Retinex model \cite{zhou_pixel_2024} & Pixel distribution remapping (PDR), pre-compensation for attenuated channels, variational model with noise and texture priors & Particle scattering, light absorption, noise \\
      \bottomrule
    \end{tabular}
  \end{adjustbox}
\end{table}

\subsubsection{Image Restoration}
Methods in this category aim to recover the original image by removing distortions caused by light scattering, absorption, and noise \cite{song_shallow_2023}. \cite{zhang_underwater_2024} introduces a Joint distortion localization and restoration model based on Progressive Guidance for underwater images.

\textbf{Denoising and Deblurring:}
Notable denoising works include Peng et al.\cite{peng_single_2015} that developed a method to compute depth maps for underwater image restoration by combining image blurriness with the image formation model (IFM) to estimate object distance. A pattern extraction-based classification system removes noise with Laplacian Bellman filtering, enhancing images using histogram equalization, and extracts target patterns using a likelihood gradient technique \cite{rajasekar_pattern_2020}. Polarimetric imaging recovery has shown potential in mitigating scattering-induced blurring and color distortion by leveraging crosstalk compensation between RGB channels \cite{liu_polarimetric_2020}. Chen et al. \cite{chen_underwater_2024} proposed the NR+FAGR framework to address class imbalance and label noise in UOD. The Noise Removal (NR) algorithm improves dataset quality by eliminating label noise, while the Factor-Agnostic Gradient Re-weighting (FAGR) algorithm rebalances gradients to treat all classes equally, leading to state-of-the-art performance in detection tasks. 

\textbf{Dehazing and Lighting Correction:}
Boffety et al. \cite{boffety_color_2012} developed a simulation tool for color restoration in underwater optical images, studying the impact of spectral discretization on color rendering and improves image color using RGB data from the simulation scene. Li et al. \cite{li_single_2016} select background light using maximally different pixels, which dehazes blue-green channels and corrects the red channel. Image restoration method in Cao et al. \cite{cao_underwater_2018} uses neural networks to estimate scene depth and backlight to address color distortion and low contrast caused by light scattering and absorption, with effectiveness confirmed by experiments. Fu et al., \cite{fu_image_2020} proposed hardware polarization-based underwater image restoration method which estimates backlight using an automatic map and introduces a new color restoration strategy for absorption, accounting for wavelength and scene depth. Wang et al., \cite{wang_range-intensity-profile_2021} proposed a method to dehaze underwater images using a single-gated image and  \cite{shao_gplm_2025} proposed a global pyramid linear modulation method that enables fine-grained feature maps. 

\subsubsection{Fusion of Restoration and Enhancement:}
Some studies have focused on both restoring and enhancing underwater images instead of addressing them separately. For example, Luo et al. \cite{luo_underwater_2021} proposed a technique for underwater image restoration using contrast optimization, color balancing, and histogram stretching. Color balancing adjusts R, G, and B channels, while histogram stretching enhances contrast and brightness using the red channel hence improving image quality and contrast. Zhou et al. \cite{zhou_adaptive_2021} address detail loss and color deviation by applying pixel-based color restoration method, histogram enhancement on the H channel, and edge preservation for image enhancement, effectively improving image quality and accuracy. Kang et al. \cite{kang_perception-aware_2023} introduced a perception-aware decomposition and fusion framework that combines two complementary pre-processed images in an independent perception-aware image space, breaking them down into mean intensity, contrast, and structure components. A hybrid contrastive learning method was proposed by Zhou et al. \cite{zhou_hclr-net_2024} to improve model robustness and generalization using non-paired positive and negative samples.  Yang and Wang \cite{yang_underwater_2025} proposed an underwater image enhancement algorithm based on multi-scale layer decomposition and fusion, which effectively improves image clarity.
Despite their progress, these methods face limitations. Enhancement may introduce artifacts that degrade object detection performance, while restoration methods often struggle with diverse and dynamic underwater conditions.

\subsection{Image Synthesis}
Image synthesis is the process of generating artificial images that resemble real-world scenes, often used to augment datasets in scenarios where collecting real data is challenging. In the underwater domain, image synthesis plays a crucial role in addressing the scarcity of annotated data, enabling the training and evaluation of object detection models in diverse and complex underwater environments. Recent advancements in generative AI, such as Variational Autoencoders (VAEs) \cite{kingma_introduction_2019}, Generative Adversarial Networks (GANs) \cite{cohen_generative_2022}, and Diffusion Models \cite{dhariwal_diffusion_2021}, have significantly enhanced the capabilities of image synthesis. Each of these approaches offers unique advantages as listed in Table.~\ref{tab:gen_AI}.

\begin{table}[h]
  \centering
  \caption{A Comparison Summary of Generative AI Models}
  \label{tab:gen_AI}
  \begin{adjustbox}{max width=\textwidth}
  \renewcommand{\arraystretch}{1.2}
    \begin{tabular}{>{\raggedright\arraybackslash}m{0.25\textwidth} >{\raggedright\arraybackslash}m{0.25\textwidth} >{\raggedright\arraybackslash}m{0.25\textwidth} >{\raggedright\arraybackslash}m{0.25\textwidth}}
      \toprule
      \textbf{Feature} & \textbf{VAEs} & \textbf{GANs} & \textbf{Diffusion Models} \\
      \midrule
      How It Works & Learns a smooth latent space; encodes and decodes data & Generator creates fake data; discriminator distinguishes real vs. fake & Iteratively adds and removes noise to generate data \\
      Image Quality & Blurry & Sharp & Realistic, high-quality, and diverse \\
      Training Stability & Stable & Unstable (mode collapse) & Stable but slow \\
      Latent Space Control & Good & Poor & Moderate \\
      Computational Cost & Low & Medium & High \\
      \bottomrule
    \end{tabular}
  \end{adjustbox}
\end{table}

In underwater applications, generative AI has been leveraged to create synthetic datasets that mimic real-world conditions. For instance, POSEIDON \cite{ruiz-ponce_poseidon_2023} employs data augmentation techniques to generate new samples by combining objects with metadata, improving data balance and model performance. Similarly, UWCNN \cite{li_underwater_2020} leverages underwater scene priors to directly reconstruct clear images without estimating imaging model parameters, synthesizing diverse underwater image degradation datasets using an underwater imaging physical model. This approach enables the training of lightweight CNNs tailored to specific water types and degradation levels, demonstrating strong generalization across various underwater scenes. Another notable method, proposed by Wu et al. \cite{wu_novel_2021} introduces an end-to-end underwater image synthesis framework that converts natural light images into synthetic underwater images using a pixel-level self-supervised training strategy, maximizing structural similarity between synthesized and real underwater images. Physics-inspired methods have also gained traction in underwater image synthesis. For example, Kaneko et al. \cite{kaneko_physics-inspired_2024} developed the PHISWID dataset, which pairs atmospheric RGB-D images with synthetically degraded underwater images by simulating color degradation and marine snow artifacts.

\subsection{Object Detection Methods}
Detection models are computational frameworks designed to identify and localize objects within images, playing a crucial role in addressing underwater challenges like low visibility, noise, and distortion. These models range from traditional methods to modern deep learning-based, transformer-based and hybrid approaches. Traditional methods rely on handcrafted features, offering simplicity and low computational cost but struggling with noise and complex underwater environments. Deep learning methods, including region-based CNNs (e.g., Faster R-CNN) with region-based models providing high accuracy and one-stage detectors (e.g., YOLO, SSD) excelling in real-time performance, extract hierarchical features for robust detection.

\begin{figure}[h]
  \centering
  \includegraphics[width=0.75\linewidth]{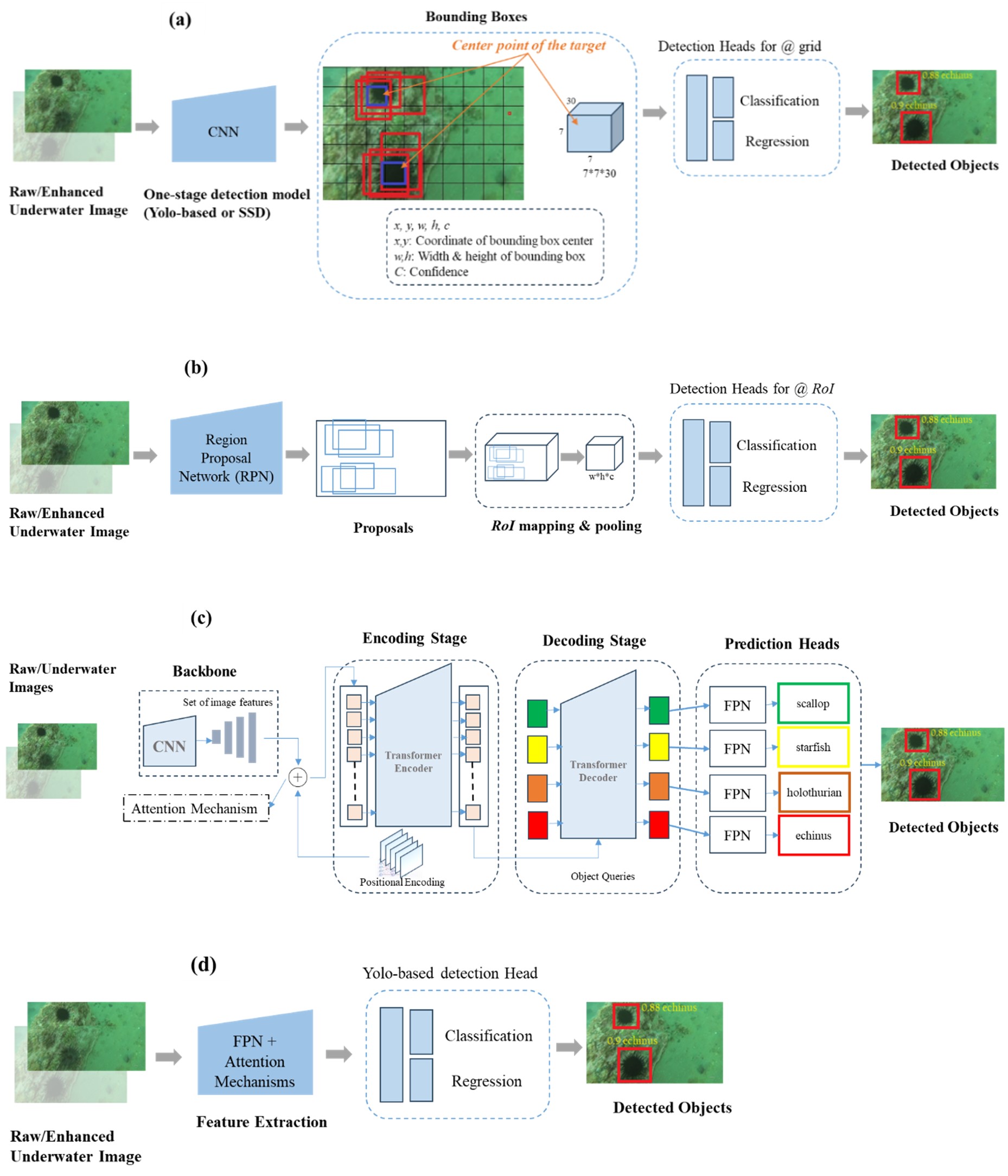}
  \caption{The Pipelines of Various Object Detection Model Architectures}
  \label{fig:model_pipelines}
\end{figure}

Transformer-based methods, like DETR, leverage self-attention mechanisms to capture global dependencies, effectively handling overlapping objects and varying scales, making them highly suitable for complex underwater scenes. Hybrid detection models combine techniques from different object detection architectures, such as integrating one-stage, two-stage, or transformer-based methods, to leverage their strengths and improve overall detection performance.
Fig.~\ref{fig:model_pipelines}:(a) one-stage detection models, represented by YOLO, which directly predict bounding boxes and class probabilities; (b) two-stage models, represented by R-CNN, which generate region proposals via a Region Proposal Network (RPN) and refine them for classification and localization; (c) transformer-based models, represented by DeTR, which utilize a CNN backbone, transformer encoders and decoders, multi-head attention mechanisms, object queries and bipartite matching for end-to-end object detection; and (d) hybrid models, combining different techniques.

\subsubsection{Techniques for Small Object Detection}
To tackle small object detection, Zhang et al. \cite{zhang_lightweight_2021}proposed a lightweight UOD method using YOLOv4 and Multi-Scale Attentional Feature Fusion to address challenges like low visibility and small targets, which show a strong balance between accuracy and speed. Chen et al. \cite{chen_novel_2022} introduced a dynamic feature fusion algorithm in the YOLOX-DFF model, which uses spatial- and scale-aware attention to combine multi-scale features and a task-aware attention module to resolve classification-regression conflicts. He et al. \cite{he_improved_2023} improved YOLOv4 by optimizing anchor sizes, adding a Squeeze-and-Excitation attention mechanism, incorporating a transformer block, and using a cascade network for better IoU matching. The YOLOv7-CHS model incorporates the High-Order Spatial Interaction (HOSI) module, which reduces model size while maintaining accuracy, and the Contextual Transformer (CT) module, which improves small target detection by combining contextual features and a simple Parameter-Free Attention (SPFA) module enhances attention \cite{zhao_yolov7-chs_2023}. Dynamic YOLO by Chen et al. \cite{chen_dynamic_2024} uses a lightweight Deformable Convolution v3 backbone for better detection, and a unified feature fusion framework that combines attention mechanisms for multi-scale features. Chen and Guo \cite{chen_sfdet_2024} proposed SFDet which uses an attention mechanism that shifts focus to the frequency domain, improving small object detection. A fusion mechanism integrates image enhancement networks for better semantic representation. PE-Transformer uses a local path detection scheme for better feature interaction and a point-based module for improved target coverage, with a weighted loss function for optimizing feature convergence \cite{gao_pe-transformer_2024}. Edge-Guided Representation Learning Network (ERL-Net) uses edge cues and an edge-guided attention module for better feature extraction, together with a hierarchical aggregation module and wide receptive field block to capture multi-scale features and small object details \cite{dai_edgeguided_2024}.

\subsubsection{Addressing Image Quality Challenges}
Significant research efforts have been dedicated to addressing image quality issues in underwater environments, which include solutions for deblurring, low contrast, occlusions, dynamic scenes, color correction, and denoising, as demonstrated in Table.~\ref{tab:address_image_quality}.

\begin{table}[h]
  \centering
  \caption{Overview of Innovative Techniques and Models Addressing Image Quality Challenges in Underwater Object Detection}
  \label{tab:address_image_quality}
  \begin{adjustbox}{max width=\textwidth}
  \renewcommand{\arraystretch}{1.2} % Increased row spacing
    \begin{tabular}{
      >{\raggedright\arraybackslash}m{0.25\textwidth} 
      >{\raggedright\arraybackslash}m{0.50\textwidth} 
      >{\raggedright\arraybackslash}m{0.25\textwidth}}
      \toprule
      \textbf{Model} & \textbf{Key Techniques/Modules} & \textbf{Addressed Challenges} \\
      \midrule
      MIPAM integrated into YOLO \cite{shen_multiple_2024} & Multiple Information Perception-based Attention Module (MIPAM): spatial downsampling, channel splitting, adaptive feature fusion & Poor imaging quality, harsh environments, concealed targets \\
      Boosting R-CNN \cite{song_boosting_2022} & RetinaRPN for high-quality proposals, probabilistic inference pipeline, boosting reweighting for hard example mining & Unbalanced light conditions, low contrast, occlusion, mimicry of aquatic organisms \\
      Modified EfficientDet \cite{jain_deepseanet_2024} & BiSkFPN (BiFPN neck with skip connections), adversarial learning, class activation maps (CAM) & Limited visibility, saline water with impurities, adversarial noise \\
      Image enhancement and multi-branch network \cite{cui_underwater_2024} & Parameter-free color correction attention module, attention-based backbone and neck, decoupled head & Color bias, haze-like effect, small objects, occlusions \\
      GCC-Net \cite{dai_gated_2024} & Real-time UIE, cross-domain feature interaction module, gated feature fusion module & Low contrast, low-light conditions \\
      CEH-YOLO \cite{feng_ceh-yolo_2024} & High-order deformable attention (HDA) module, ESPPF module, composite detection (CD) module, WIoU v3 loss & Low contrast, color variations, small/overlapping objects \\
      YOLOv8m \cite{bajpai_enhancing_2024} & State-of-the-art deep learning architecture & Limited visibility, inconsistent lighting \\
      HLASwin-T-ACoat-Net \cite{manimurugan_hlaswin-t-acoat-net_2024} & Hybrid Local Acuity Swin Transformer, Adapted Coat-Net, CLAHE, path aggregation network, ROI alignment & Limited visibility, scattering, absorption, false detection \\
      \bottomrule
    \end{tabular}
  \end{adjustbox}
\end{table}

\subsubsection{Enhancing Bounding Box Accuracy in Object Detection}
Bounding box inaccuracies arise when predicted boxes fail to align precisely with actual objects, often due to irregular object shapes, occlusions, and degraded image quality. Chen et al. \cite{chen_bounding_2020} introduced a bounding box repairing algorithm which enhanced the mask Scoring R-CNN network by refining coarse detection results. The algorithm calculates Intersection over Union (IoU) to match and repair bounding boxes, demonstrating significant improvements in detection and localization.

\subsubsection{Handling Scale Variations in Underwater Object Detection}
\begin{figure}[h]
  \centering
  \includegraphics[width=0.6\linewidth]{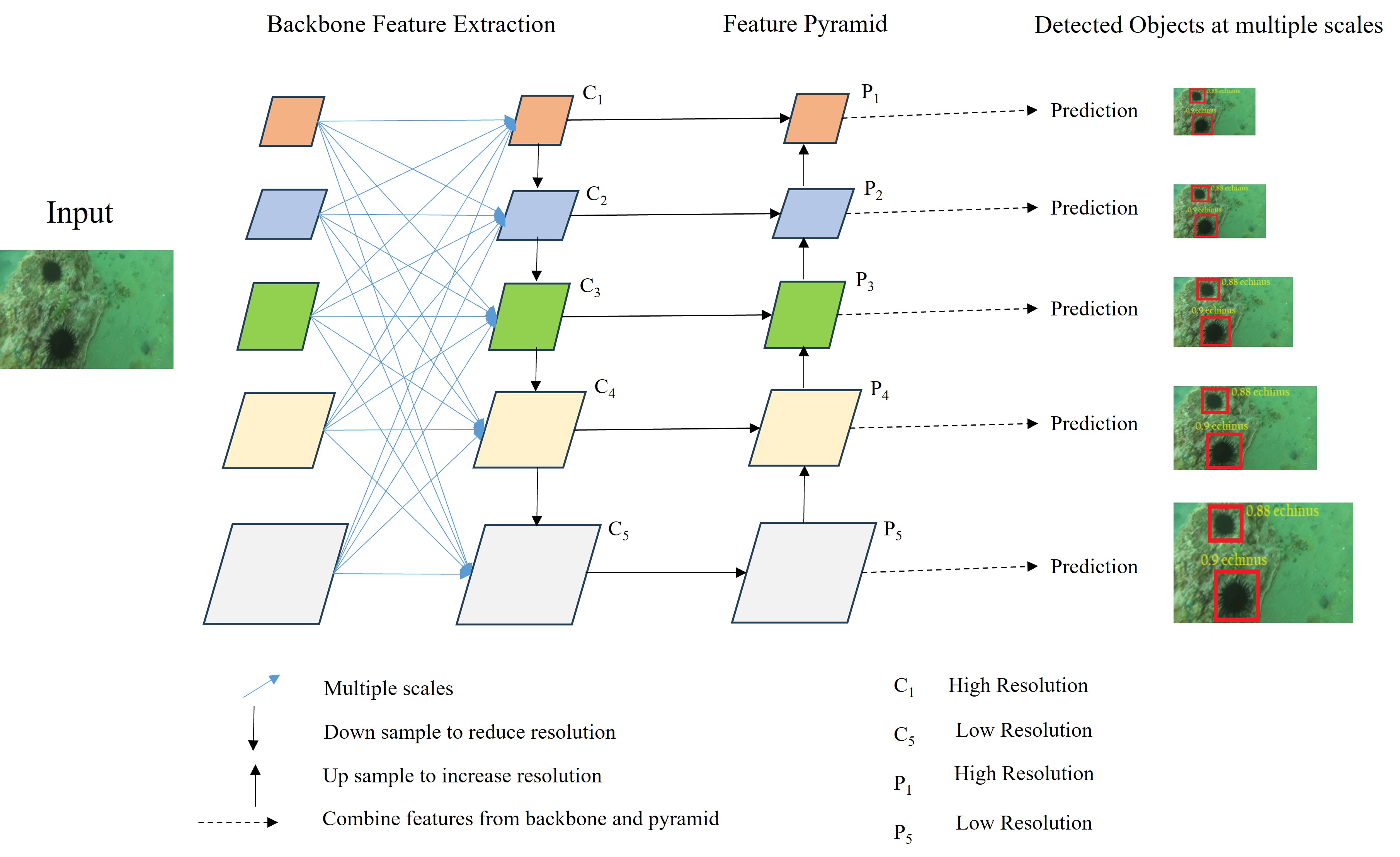}
  \caption{Multi-Scale Feature Extraction and Fusion in Object Detection}
  \label{fig:multiscale}
\end{figure}

Scale variations occur when objects appear at different sizes due to varying distances from the camera or changes in perspective. For this case, Xu et al. \cite{xu_scale-aware_2021} introduced a scale-aware feature pyramid architecture (SA-FPN) that uses a specialized backbone for better feature extraction and a multi-scale feature pyramid with context information. Soft non-maximum suppression reduces duplicate bounding boxes for improved accuracy. Shi et al. \cite{shi_underwater_2021} improved the Faster-RCNN algorithm by replacing the original backbone with ResNet for better feature extraction and adds BiFPN for multi-scale feature fusion, while EIoU reduces redundant bounding boxes, and K-means++ clustering optimizes anchor boxes for detecting objects with varying shapes and sizes. Fig.~\ref{fig:multiscale}., illustrates the multi-scale feature extraction and fusion process, highlighting how downsampling, upsampling, and lateral connections are utilized to enhance object detection across varying resolutions. The input image undergoes backbone feature extraction, where different resolution feature maps \(C_1\) - \(C_5\) are generated. These features are then processed through a feature pyramid network (FPN) to enhance detection across multiple scales \(P_1\) - \(P_5\). The final predictions show detected objects at different resolutions, improving accuracy for small and occluded underwater targets. Ling et al. \cite{ling_underwater_2023} enhanced the YOLOv7-tiny model for UOD by integrating the CBAM for better feature extraction, Soft-NMS for improved accuracy with overlapping objects, and CARAFE for better multi-scale feature fusion. These improvements led to significantly higher detection accuracy compared to YOLOv7-tiny, YOLOv5, and YOLOv8. Wang and Zhao \cite{wang_improved_2024} proposed an improved YOLOv8-MSS algorithm which includes a small target detection head for better sensitivity, the C2f\_MLCA module for noise robustness, and SENetV2 to strengthen anti-interference capabilities. The SIoU loss function improves accuracy by considering shape and geometry. Feature Boosting and Differential Pyramid Network (FBDPN) by Ji et al. \cite{ji_fbdpn_2024} combines CNN and Transformer strategies with an FPN-inspired architecture for better multi-scale feature learning and long-distance dependencies, where modules like NSFBM enhance context, while CSFDM reduces scale redundancy. Lin et al. \cite{lin_underwater_2024} improved DETR by incorporating a learnable query recall mechanism for reducing noise, and a lightweight adapter for enhancing multi-scale feature detection.

\subsubsection{Real-Time Detection}
Real-time detection is essential for applications such as AUVs and ROVs, where quick and accurate responses are crucial. In relation to this, Wang et al. \cite{wang_self-supervised_2023} introduced LUO-YOLOX, a lightweight detection network which reduces computational complexity by using a weighted ghost-CSPDarknet and simplified PANet, while a self-supervised pre-training framework (UAET) enhances feature extraction from degraded images. Lee and Chen \cite{lee_artificial_2023} implemented a Siamese Region Proposal Network with shared weights to track moving targets and handle one-shot detection for unidentified objects at a tracking rate of up to 180 FPS. Lightweight TC-YOLO uses adaptive histogram equalization for image enhancement, transformer self-attention in the backbone, and coordinate attention in the neck for better feature extraction \cite{liu_underwater_2023}. Ding et al. \cite{ding_lightweight_2024} introduced PDSC-YOLOv8n, a lightweight network to address misidentification, missed detection, and low visibility. Using enhanced data augmentation, Ghost and GSConv modules for a lightweight structure, and attention mechanisms like GAM and CBAM, the model improves target feature extraction and generalization. Zhang et al. \cite{zhang_efficient_2024} optimized YOLOv8 by using FasterNet-T0 and adding a small-object prediction head, reducing model size by over 22\%. Deformable Convolutional Networks and Coordinate Attention improved detection of irregularly shaped targets. A lightweight YOLOv8-based UOD algorithm incorporates an Adaptive Underwater Image Enhancement module for better image quality and a Re-parameterized Partial Convolution Block to reduce network parameters, demonstrating model balance in accuracy and speed \cite{zhao_lightweight_2024}. Pan et al. \cite{pan_optimization_2024} improved YOLOv9s-UI by using a Dual Dynamic Token Mixer (D-Mixer) from TransXNet for better feature extraction and a feature fusion design with channel and spatial attention mechanisms, improving accuracy while keeping the model compact (9.3M parameters). EnYOLO by Wen et al. \cite{wen_enyolo_2024}implements a shared lightweight backbone, reducing latency and computational costs. \cite{du_cluster-based_2024} proposes an energy-efficient underwater target detection scheme using soft and hard decision fusion.

\subsubsection{Addressing Class Imbalance in UOD}
Class imbalance in UOD arises when certain object classes dominate the dataset, resulting in biased models that struggle to detect minority classes. To solve this, Wang and Xiao. \cite{wang_underwater_2023} improved Faster R-CNN for underwater detection by replacing the VGG16 backbone with Res2Net101, using OHEM to address class imbalance, and optimizing bounding box regression with GIOU and Soft-NMS. Chen et al. \cite{chen_underwater_2024} address class imbalance and label noise in UOD by proposing a noise removal (NR) algorithm to eliminate label noise and a factor-agnostic gradient re-weighting algorithm (FAGR) to rebalance gradients across classes, ensuring equal treatment of all classes and minimizing detection discrepancies, achieving state-of-the-art performance on underwater datasets.

\needspace{2\baselineskip}
\subsubsection{Domain Shift in UOD}
As mentioned earlier, this issue occurs when the environmental conditions of the training dataset, such as lighting, visibility, and depth, differ significantly from those of the target deployment environment which can lead to reduced model performance and generalization. To address domain shift, Liu et al. \cite{liu_towards_2020} introduced GUOD to address domain shift in underwater detection, using the WQT data augmentation method for diverse training samples. The DG-YOLO model, combining YOLOv3 with a Domain Invariant Module, enhances feature extraction, and experiments show improved domain generalization in varied underwater environments. A domain generalization framework \cite{chen_achieving_2023} Domain Mixup and Contrastive Learning to generate diverse training domains through style transfer and enhance robustness. Contrastive loss ensures domain-invariant feature learning while maintaining discriminative power. NFD-YOLO enhances YOLOv5 with a feature manipulation module and noise-agnostic subnetworks to focus on relevant features. The ACmix attention module improves learning for small datasets improving generalization and robustness \cite{yu_treat_2023}. While significant progress has been made in UOD, achieving robust detection remains a challenge. LVLMs offers a promising direction to address these persistent issues. LVLMs' ability to integrate multimodal data, generalize across domains, and leverage contextual relationships positions them as a strong candidate for overcoming current limitations. Their success in other vision-related tasks underscores their potential to enhance accuracy, robustness, and scalability in complex underwater environments.\\

\section{ROLE OF LVLMS}
LVLMs' ability to integrate multimodal data, generalize across domains, and leverage contextual relationships makes them a strong candidate for overcoming existing limitations in UOD.

\begin{table}[ht]
  \centering
  \caption{Overview of some recent LVLMs and their capabilities, highlighting key examples, their developers, release time, and specialized tasks. VQA: Visual Question Answering, REC: Referring Expression Comprehension, VCR: Visual Commonsense Reasoning, DQA: Document Question Answering \cite{caffagni_revolution_2024}}
  \label{tab:recent_LVLMs}
  \begin{adjustbox}{max width=\textwidth}
  \renewcommand{\arraystretch}{1.2}
    \begin{tabular}{>{\raggedright\arraybackslash}m{0.25\textwidth} >{\raggedright\arraybackslash}m{0.1\textwidth} >{\raggedright\arraybackslash}m{0.65\textwidth}}
      \toprule
      \textbf{Model} & \textbf{Year} & \textbf{Main Tasks and Capabilities} \\
      \midrule
      VisualBERT \cite{li_visualbert_2019} & 2019 & VQA, Captioning, REC, Visual Grounding \\
      CLIP \cite{radford_learning_2021} & 2021 & VQA, Object Detection, VCR, Visual Grounding, Retrieval \\
      Flamingo \cite{alayrac_flamingo_2022} & 2022 & Visual Dialogue, VQA, Captioning \\
      GPT-4V \cite{yang_dawn_2023} & 2023 & VQA, Texts-in-Image tasks, Visual understanding and analysis, Visual creative and debugging \\
      BLIP-2 \cite{li_blip-2_2023} & 2023 & Visual Dialogue, VQA, Captioning, Retrieval \\
      Gemini \cite{gemini_team_gemini_2023} & 2023 & VQA, Captioning, REC, Object Detection, Speech Generation, Understanding, Code Writing, Debugging, Mathematical Reasoning, Multilingual \\
      PaLM-E \cite{driess_palm-e_2023} & 2023 & Captioning, VQA, Object Detection, Visual Grounding, Image Classification \\
      MiniGPT-v2 \cite{chen_minigpt-v2_2023} & 2023 & Visual Dialogue, VQA, Captioning, Referring, REC, Visual Grounding \\
      InternVL \cite{chen_internvl_2023} & 2023 & Visual Dialogue, VQA, Captioning \\
      Qwen2-VL \cite{bai_qwen-vl_2023} & 2024 & Visual Dialogue, Multilingual, VQA, Captioning, REC \\
      LLaMA 3.2-vision \cite{grattafiori_llama_2024} & 2024 & Visual Dialogue, Multilingual, VQA, Captioning, DQA, Visual Grounding, Image Understanding, Analysis \\
      Baichuan OM \cite{li_baichuan-omni_2024} & 2024 & Visual Dialogue, Image Understanding, Text-to-Speech \\
      TransFusion \cite{zhou_transfusion_2024} & 2024 & Object Detection, LiDAR-Camera Data Fusion, Multi-Modality Feature Fusion \\
      Molmo \cite{deitke_molmo_2024} & 2024 & Clock Face Recognition, VQA, Counting \\
      Florence-2 \cite{xiao_florence-2_2023} & 2024 & Visual Dialogue, VQA, Captioning, Object Detection \\
      DeepSeekVL2 \cite{wu_deepseek-vl2_2024} & 2024 & VQA, Object Detection, Segmentation, Facial Recognition, OCR, Image Enhancement, Video Analysis, Scene Understanding, 3D Reconstruction \\
      \bottomrule
    \end{tabular}
  \end{adjustbox}
\end{table}

\subsection{Evolution of LVLMs }
LVLMs represent a natural progression from Large Language Models (LLMs), which provide the linguistic foundation for these multimodal systems. LLMs, such as GPT-4 \cite{openai_gpt-4_2023}, Gemini \cite{gemini_team_gemini_2023}, Palm 2 \cite{anil_palm_2023}, DeepSeek \cite{wu_deepseek-vl2_2024}, LLaMA \cite{grattafiori_llama_2024}, excel in natural language understanding, reasoning, and generation tasks. These models are trained on massive corpora of textual data, allowing them to generalize across diverse linguistic tasks. By extending the capabilities of LLMs to incorporate visual data, LVLMs enable complex reasoning and contextual understanding that bridge text and image inputs. 
As of January 2025, several prominent LVLMs have emerged, each bringing unique innovations to the field. Table.~\ref{tab:recent_LVLMs}., highlights some of these recently developed LVLMs and their capabilities. Given their transformative ability to integrate visual and textual reasoning, their application to UOD holds significant promise. The adaptability and success of models like EarthGPT \cite{zhang_earthgpt_2024} in remote sensing object detection suggest that LVLMs can be optimized for underwater tasks, bridging current gaps in performance and generalization.

\subsection{Potential of LVLMs in Addressing UOD Challenges }
LVLMs offer innovative solutions to persistent challenges in visual tasks, demonstrating their capability to enhance various applications through multimodal learning.

\subsubsection{Enhancing Image Quality}
Models like the LLMRA \cite{jin_llmra_2024} leverage LVLMs to enhance low-quality images by extracting degradation information and encoding it as contextual embeddings. This process involves generating textual descriptions of the image's degradation attributes, which are integrated into the restoration network through specialized modules like the Context Enhance Module (CEM) and the Degradation Context-based Transformer Network (DC-former), addressing issues like blurring, noise, and low contrast. In underwater environments, where images are heavily degraded, LVLMs can similarly exploit their contextual understanding and extensive pretraining to identify and correct degradation patterns.

\subsubsection{Advancing Target Detection}
ViGoR \cite{yan_vigor_2024} enhance visual grounding through fine-grained reward modeling, improving the accuracy of object localization and attribute recognition. Similarly, RelationVLM \cite{huang_relationvlm_2024} uses multi-stage relation-aware training to improve detection in complex scenes with overlapping and small objects, making it suitable for high-density environments such as underwater.

\subsubsection{Solving Dataset Scarcity and Diversity Challenges}
The approach presented in CoLLaVO \cite{lee_collavo_2024} offers an innovative solution, by leveraging instruction tuning and a visual prompt tuning scheme (Crayon Prompt) based on panoptic color maps, CoLLaVO enhances object-level image understanding and improves zero-shot performance on vision-language tasks. The model’s learning strategy, Dual QLoRA, preserves and fine-tunes object-level understanding without forgetting previously learned features. For UOD, where data scarcity and domain shifts are prevalent, CoLLaVO’s ability to excel in zero-shot settings can help mitigate reliance on large annotated datasets. By generating a versatile understanding of visual objects, the model can generalize better across underrepresented classes and unseen underwater environments, improving detection accuracy even when labelled data is limited. Additionally, synthetic data generation using DALL-E has shown exceptional promise in addressing data insufficiency and class imbalance. DALL-E's ability to create high-quality, context-specific synthetic images enables the generation of labeled datasets tailored to specific domains. In the medical domain \cite{adams_what_2023}, DALL-E 2 has been employed to generate synthetic radiological images, helping to augment datasets for rare disease detection while maintaining image quality and relevance to real-world conditions. Similarly, in agriculture \cite{sapkota_creating_2023}, DALL-E has been utilized to produce realistic images of crop diseases across various conditions. Expanding on these advancements, DALL-M \cite{hsieh_dall-m_2024} integrates generative AI with domain-specific knowledge to effectively synthesize data aligned with clinical scenarios, improving dataset diversity for medical imaging tasks such as tumor and fracture detection.

LLM-synthesized data is a powerful augmentation technique for UOD. By crafting precise prompts, it enables control over object types, quantities, locations, sizes, colors, and camouflage, while addressing class imbalance. Unlike traditional methods like flipping or rotation, which can distort objects and reduce realism.

\subsubsection{Streamlining Computation for Real-time Applications}
The MoE-Tuning strategy proposed in Lin et al. \cite{lin_moe-llava_2024}, along with the MoE-LLaVA architecture, offers a significant advancement in addressing the computational and processing challenges by employing a sparse model architecture, where only the top-k experts are activated during deployment, while the remaining parameters remain inactive. This reduces the computational burden without compromising performance, notably achieving comparable results with a much smaller number of activated parameters. For UOD, this approach could be highly beneficial for real-time deployment on AUVs and UAVs. The InternLM-XComposer2-4KHD method in \cite{dong_internlm-xcomposer2-4khd_2024} offers a novel solution to address this issue by optimizing the balance between resolution and computational efficiency. This method introduces a dynamic resolution paradigm with automatic patch configuration, which allows the model to adaptively handle image resolutions ranging from 336 pixels to 4K HD (3840 x 1600). This strategy ensures that the computational cost remains manageable, as the resolution can be dynamically adjusted based on the specific input or hardware constraints. In UOD, it ensures that critical object details, often lost in degraded underwater environments, are effectively captured and analysed.

\subsubsection{Enhancing Object Detection Accuracy and Localization}
The Positional Insert (PIN) method introduced in \cite{dorkenwald_pin_2024} tackles issues related to localization and bounding box predictions by leveraging frozen Vision-Language Models (VLMs), such as Flamingo or GPT-4V, which are typically trained on caption-based multimodal data but lack explicit spatial grounding for localization. Instead of requiring additional bounding box annotations or supervised detection data, the PIN module introduces a learnable spatial prompt that enables object localization within a VLM. This spatial prompt is lightweight, input-agnostic, and trained through a next-token prediction task on synthetic data. By sliding the PIN module inside the frozen VLM, the method unlocks the model's object localization capabilities without adding new detection-specific output heads, ensuring scalability. 

In case of UOD, this approach can be transformative in addressing localization challenges and zero-shot localization without extensive annotated detection datasets. 
LVLMs offer great potential for UOD but may face challenges like high computational demands, difficulty adapting to underwater conditions, and limited annotated data. Issues like noise in underwater images and model hallucinations also need addressing. Despite these challenges, advancements in hardware, synthetic data and Efficient fine-tuning can help overcome these limitations, making LVLMs a promising tool for underwater applications. In the next section, two case studies explore how synthetic data generation by LVLMs and efficient fine-tuning strategies can help mitigate underlying challenges in UOD.

\section{CASE STUDIES/USE CASES}
In our review, we conducted two case studies to strengthen our analysis and provide practical insights. These case studies focused on underwater image synthesis, exploring the generation of synthetic underwater datasets, and fine-tuning LVLMs for UOD tasks, highlighting their potential to address domain-specific challenges.

\subsection{Leveraging Generative LVLMs for Synthetic Data Augmentation in UOD}

\begin{figure}[h]
  \centering
  \includegraphics[width=0.8\linewidth]{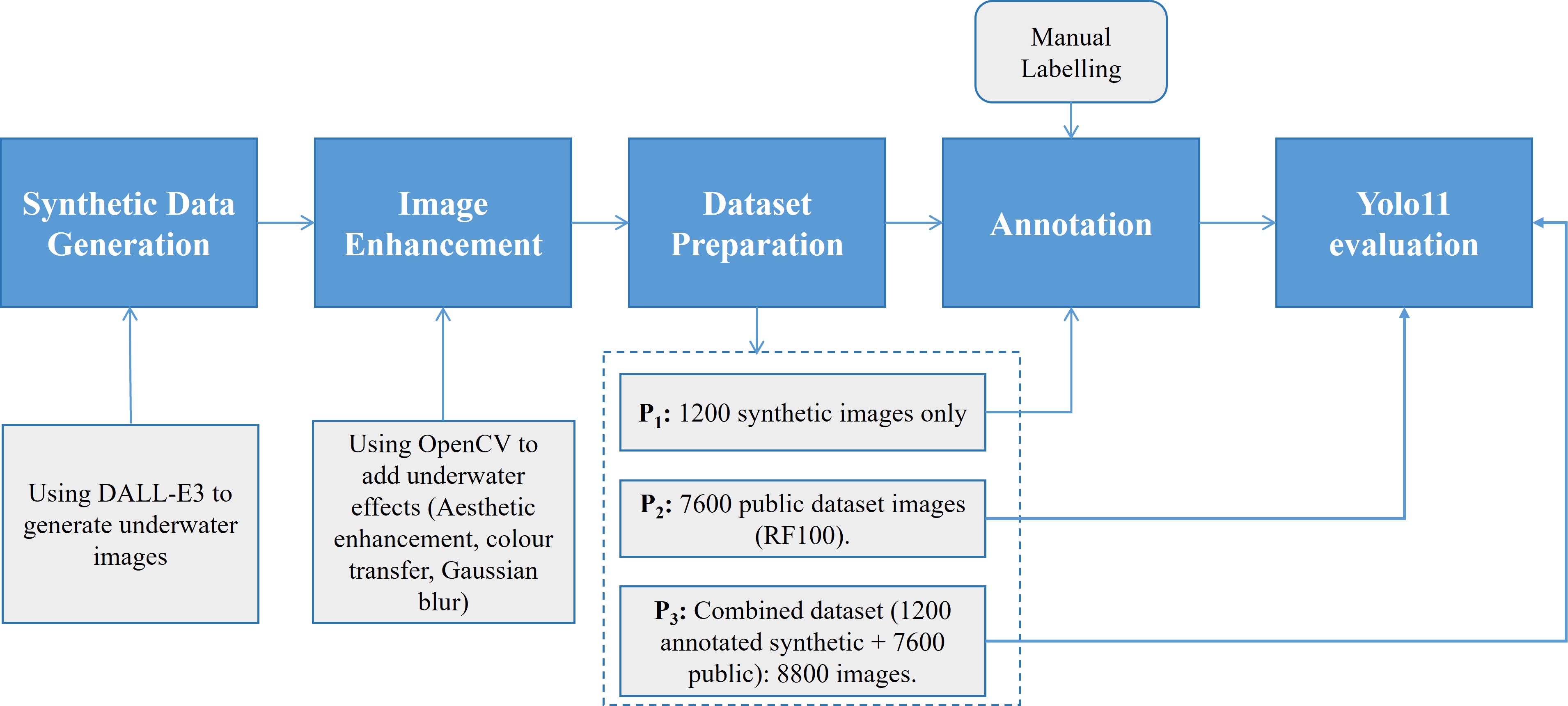}
  \caption{The Workflow Showing the effect of Synthetic data augmentation and Enhancement in Improving Underwater Object Detection evaluated on YOLO11}
  \label{fig:synthetic_data_generation}
\end{figure}

In this section, we explore the potential of generating synthetic underwater images using DALL-E 3, to augment the real 'Roboflow100' public underwater dataset. Our aim is to evaluate the potential of a LVLM synthetic data augmentation technique to improve object detection model performance in challenging UOD tasks. 
As demonstrated in Fig.~\ref{fig:synthetic_data_generation}., the pipeline presents a systematic approach to evaluating UOD through the integration of synthetic and real datasets.

The process begins with synthetic data generation, where DALL-E 3 is employed to create initial realistic underwater images. The generated images are subsequently refined through image enhancement using OpenCV, which applies underwater-specific effects such as aesthetic enhancement, color transfer and Gaussian blur. During dataset preparation, three paths are established: \( P_1 \) only synthetic data, \( P_2 \) only public data (RF100 dataset), and then \( P_3 \) a mixture of the two datasets. The synthetic dataset firstly undergoes manual annotation to ensure precise labeling before being combined and subjected to evaluation. YOLO11 is evaluated on \( P_2 \) and \( P_3 \). This pipeline highlights the potential of synthetic data augmentation and hybrid datasets to improve the robustness of UOD. 

\textbf{Text-to-Image Generation:} The first approach involved creating detailed text prompts through DALL-E 3 for image generation. These prompts were carefully designed to reflect specific underwater objects, such as scallops, sea cucumber, sea urchin and starfish, along with the corresponding environmental conditions. DALL-E 3 utilized these textual descriptions to generate a wide variety of underwater images with different object types, compositions, and backgrounds. 

\begin{figure}[h]
  \centering
  \includegraphics[width=0.8\linewidth]{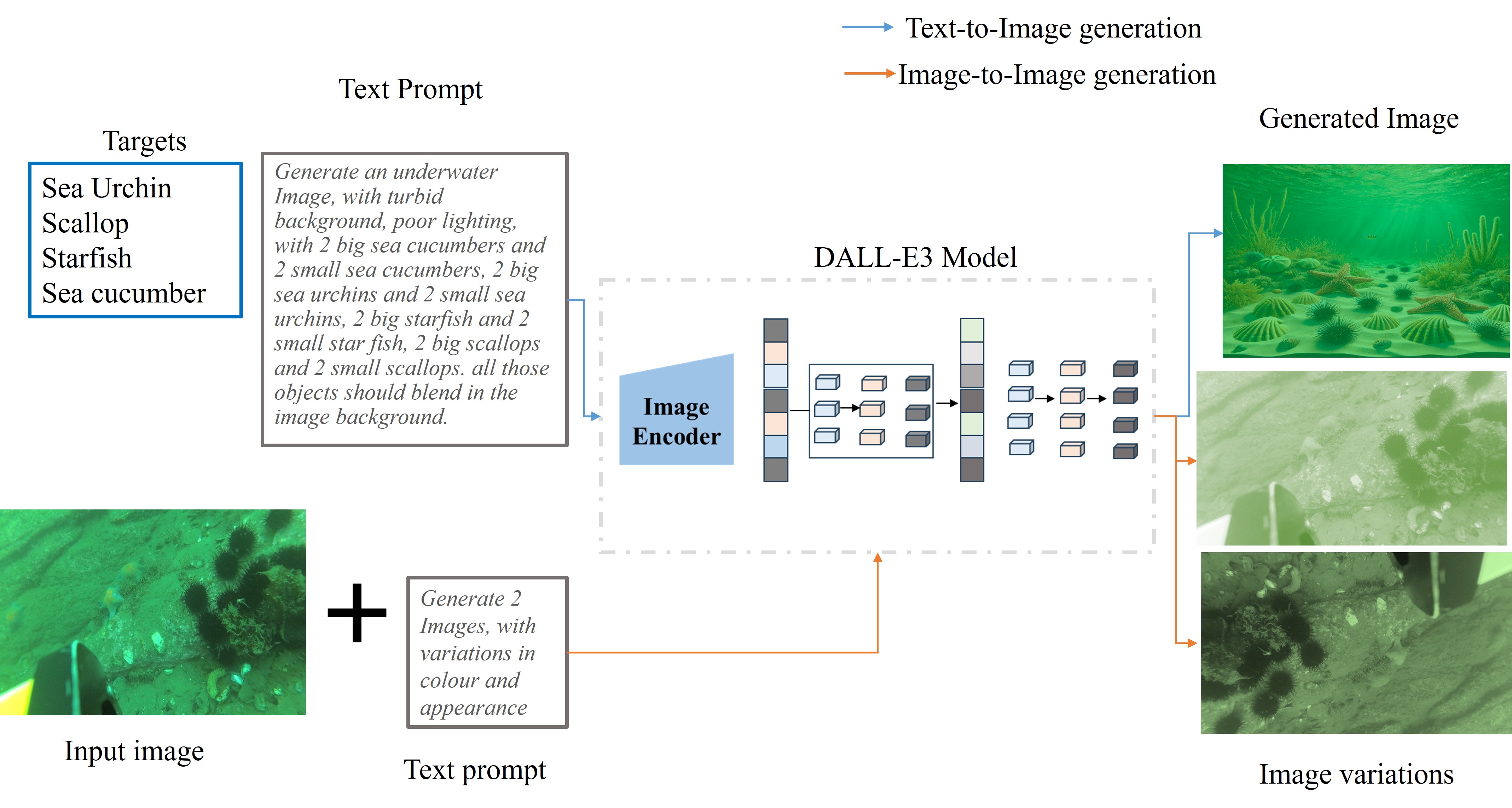}
  \caption{Synthetic Data Generation Pipeline}
  \label{fig:pipeline_synthetic}
\end{figure}

\textbf{Image-to-Image Variation:} To further increase the diversity of the synthetic dataset, an image-to-image variation technique was employed. Here, reference images from the real underwater dataset were paired with text prompts. These prompts guided DALL-E 3 to generate variations of the reference images, maintaining the core underwater object features while introducing changes in lighting, background, object positioning, and scene composition. This method was particularly useful for creating a more varied set of images that reflected the types of objects and environments typically found in real underwater scenes Fig.~\ref{fig:pipeline_synthetic}.

DALL-E 3’s text-to-image and image-to-image generation paths produced clear and high-resolution underwater images, however, the generated images appeared overly pristine and lacked the typical noise and disturbances found in real underwater scenes. To address this, image enhancement techniques were applied using OpenCV (in the next subsection) to introduce subtle underwater effects such as color adjustments and blur, ensuring the synthetic images closely resembled real underwater environments. Care was taken to avoid overly turbid enhancements, as excessively degraded images often require restoration, which was beyond the scope of this process.

\subsubsection{Image Enhancement on DALL-E 3’s Generated Images}
\begin{figure}[h]
  \centering
  \includegraphics[width=0.7\linewidth]{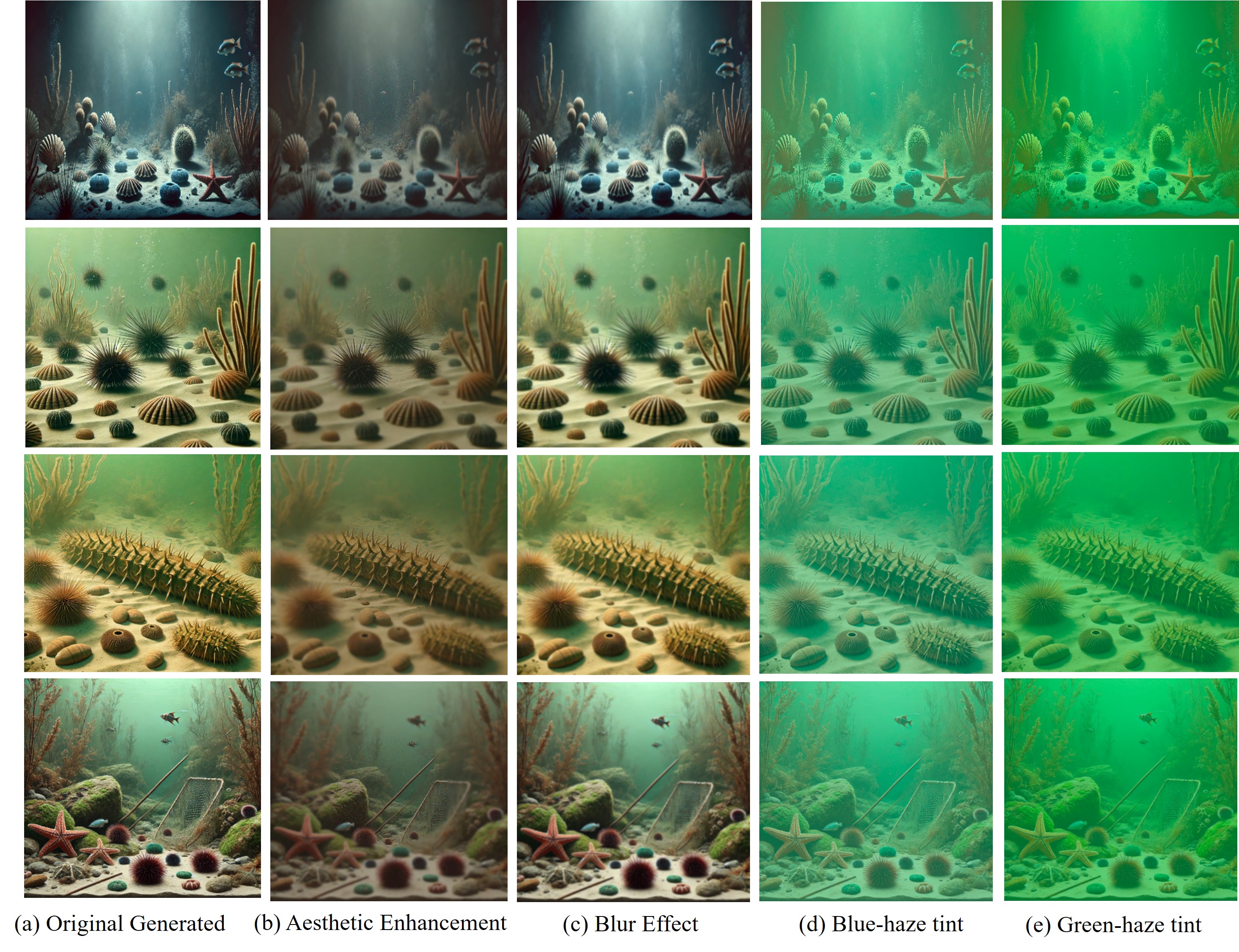}
  \caption{Sample Synthetic Generated Underwater Images with desired Target Object Classes (echinus, holothurian, scallop and starfish) with Applied Enhancement Effects, Showcasing Improved Visual Quality and Realism}
  \label{fig:generated_and_enhanced}
\end{figure}

In this work, image enhancement plays a critical role in bridging the gap between synthetic underwater images and real-world underwater scenes. OpenCV was utilized to apply subtle enhancements such as color adjustments, aesthetic transformations, and blur, ensuring the synthetic images appeared more realistic. These enhancement techniques aimed to simulate underwater conditions while preserving visual clarity suitable for object detection tasks. The 1200 synthetic images were divided into 5 groups: 240 with no any effects, 240 with aesthetic enhancement, 240 with Gaussian blur effect, 240 with blueish haze tint and other 240 images with greenish-haze tint. The detailed steps on how these enhancements were performed are provided in Appendices: A.1, A.2 and A.3. After enhancement effects, Fig.~\ref{fig:generated_and_enhanced}. showcases sample synthetic underwater images with improved visual quality and realism.

\FloatBarrier % Ensures no floats move beyond this point

\subsubsection{Dataset Annotation and Preparation}
After applying the enhancement effects, the synthetic images still lacked annotations, which are essential for object detection tasks. To address this, we manually annotated all 1,200 synthetic images using the Roboflow annotation tool, ensuring precise labeling of objects in each image. Once the annotations were completed, the synthetic dataset was combined with real underwater images to create a hybrid dataset. To evaluate the impact of our augmentation technique, we trained the YOLO11 model separately on both the combined dataset and the original underwater dataset, allowing us to compare their performance and understand the effectiveness of incorporating synthetic data in improving UOD. Table.~\ref{tab:x_tics}, shows the dataset composition and characteristics.

\begin{table}[h]
  \centering
  \caption{Dataset Composition and Characteristics}
  \label{tab:x_tics}
  \begin{adjustbox}{max width=\textwidth}
  \renewcommand{\arraystretch}{1.2}
    \begin{tabular}{>{\raggedright\arraybackslash}m{0.2\textwidth} >{\raggedright\arraybackslash}m{0.2\textwidth} >{\raggedright\arraybackslash}m{0.15\textwidth} >{\raggedright\arraybackslash}m{0.4\textwidth}}
      \toprule
      \textbf{Dataset} & \textbf{Number of Images} & \textbf{Object Classes} & \textbf{Environmental Variations} \\
      \midrule
      Original Dataset & 7600 & 4 & High turbidity, low visibility, complex scenes \\
      Synthetic Dataset & 1200 & 4 & Moderate turbidity, moderate visibility, moderate scenes \\
      Combined Dataset & 8800 & 4 & Mixed \\
      \bottomrule
    \end{tabular}
  \end{adjustbox}
\end{table}

\subsubsection{Experimental Results}
\begin{figure}[h]
  \centering
  \includegraphics[width=0.8\linewidth]{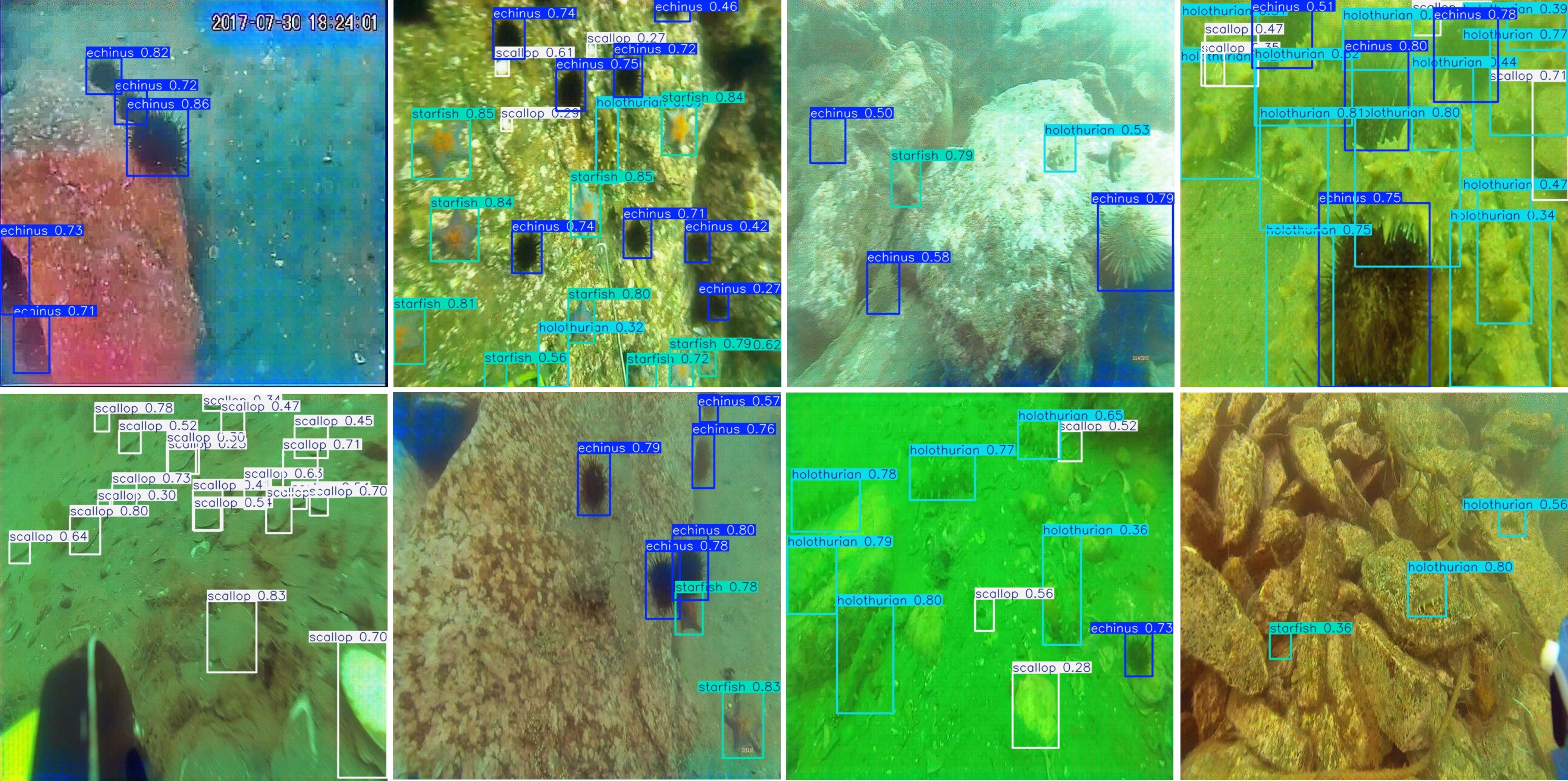}
  \caption{YOLO11 inference results on underwater images after training with a combined dataset, demonstrating detection performance in an underwater environment}
   \label{fig:yolo11_inference}
\end{figure}

The datasets were trained on the YOLO11 model (The most recent real-time yolo version) to evaluate the impact of synthetic data augmentation on UOD performance. From Table.~\ref{tab:inference_yolo11}, results demonstrate the potential benefits of integrating synthetic data into UOD.Training the YOLO11 model on the combined dataset achieved slightly higher performance metrics compared to the original dataset alone, with improvements observed in both mAP@50 (0.796 vs. 0.793) and mAP@50-95 (0.505 vs. 0.501). Additionally, the combined dataset showed improved recall (0.736 vs. 0.714), indicating a better ability to detect true positive instances, though there was a slight trade-off in precision, as indicated by a marginal drop in box precision (0.780 vs. 0.805). These results highlight the strength of synthetic data augmentation in improving the generalization and detection performance of the model, especially in capturing objects that may not be well-represented in the original dataset. YOLO11 inference results on underwater images after training with a combined dataset demonstrate great localization, classification, and regression performance in real underwater environments, as shown in Fig.~\ref{fig:yolo11_inference}.

\subsubsection{Discussions and Insights}
To evaluate the effectiveness of the synthetic data generated by LLMs, we focus on three key data features: quality, diversity, and complexity, which are critical for improving model performance in open-ended tasks such as UOD. This approach is inspired by recent research by Havrilla et al. \cite{havrilla_surveying_2024} on effects of quality, diversity and complexity in synthetic data from LLMs. Specifically, the study highlights that quality is crucial for ensuring the model’s ability to generalize within the training distribution, diversity is essential for improving out-of-distribution generalization, and complexity plays a vital role in refining model robustness and adaptability.

\begin{table}[h]
  \centering
  \caption{Evaluation of Yolo11 performance on combined and original underwater dataset}
  \label{tab:inference_yolo11}
  \begin{adjustbox}{max width=\textwidth}
  \renewcommand{\arraystretch}{1.2}
    \begin{tabular}{>{\raggedright\arraybackslash}m{0.25\textwidth} >{\raggedright\arraybackslash}m{0.15\textwidth} >{\raggedright\arraybackslash}m{0.15\textwidth} >{\raggedright\arraybackslash}m{0.15\textwidth} >{\raggedright\arraybackslash}m{0.15\textwidth}}
      \toprule
      \textbf{Dataset} & \textbf{mAP@50} & \textbf{mAP:50-95} & \textbf{Recall} & \textbf{Precision} \\
      \midrule
      Original Dataset & 0.793 & 0.501 & 0.714 & 0.805 \\
      Combined Dataset & 0.796 & 0.505 & 0.736 & 0.780 \\
      \bottomrule
    \end{tabular}
  \end{adjustbox}
\end{table}

\textbf{Quality:}
The synthetic images generated by DALL-E 3 were visually clear and of high resolution, representing a significant strength in terms of quality. In contrast to real underwater images, which are often affected by blurriness, low light, and noise due to challenging environmental conditions, the synthetic images provided a sharp and clean representation of underwater objects. Enhancements applied to these images further improved their resemblance to realistic underwater scenarios while preserving the critical features necessary for detection models. The quality of the enhanced images was rigorously evaluated using metrics such as PSNR (Peak Signal-to-Noise Ratio) and SSIM (Structural Similarity Index) to ensure their closeness to reference images, maintaining both visual fidelity and utility for downstream tasks.

\textbf{Diversity:}
The synthetic dataset exhibited significantly higher diversity than the real dataset, achieved through controlled variations in object placement, backgrounds, instances, and lighting conditions. Text-to-image and image-to-image methods enabled the creation of diverse scenes with varying objects, lighting, and water depths, enhancing out-of-distribution generalization. While this diversity augmented the more uniform 'Roboflow100' real dataset, a gap remained between synthetic and real-world variability. Real data captured unpredictable underwater conditions, such as object occlusion and water quality changes, which were not fully replicated in the synthetic dataset.

\textbf{Complexity:}
To address the issue of complexity, we intentionally introduced controlled color effects such as greenish haze, bluish tint, blur, and turbidity, using OpenCV, to simulate the environmental challenges typically found in real underwater scenarios. While the added color effects improved the resemblance to underwater conditions, the overall complexity of the dataset remained lower than that of real underwater images. Real underwater datasets often contain more noise, varying light conditions, and intricate object occlusions, which are more challenging for detection models to handle. It is worth noting, however, that the added effects did not need to create overly turbid images. After all, underwater images are often enhanced to increase visibility before training or evaluation tasks. This balance ensured that the synthetic dataset retained sufficient clarity for detection while introducing some of the key environmental challenges of underwater scenes.

\textbf{Strengths of and weaknesses of Synthetic Data Generation Using DALL-E 3}.
Synthetic data generation using DALL-E 3 offers several strengths for UOD. (i) It increases data diversity through text-to-image and image-to-image variations, enabling the creation of unique datasets. (ii) The quality of synthetic images can be controlled to include realistic underwater effects such as haze, blur, and turbidity, enhancing their utility. (iii) By carefully engineering prompts, balanced object classes can be achieved, addressing class imbalance issues. (iv) It provides efficient data augmentation, reducing dependence on re-annotated datasets that often suffer from repetitive images and limited diversity in real underwater data. However, there are notable limitations. (i) Synthetic images often lack natural imperfections like turbidity, lighting variability, and occlusions, which limits their generalization to real-world underwater conditions. (ii) Generating realistic underwater images is challenging due to the scarcity of training data for underwater domains, unlike common in-air objects that are well-represented in LVLM datasets. (iii) Synthetic images fail to fully capture complex underwater phenomena such as light scattering, water turbidity, and unpredictable occlusions. (iv) Generated images require manual annotation, which is time-consuming and resource-intensive.
The small performance gap between the combined dataset and the original dataset suggests that the limited number of synthetic images (1,200) might not have been sufficient to blend effectively with the larger real dataset (7,600). Generating a larger synthetic dataset and retraining the model could yield more significant improvements. At the same time, generating more synthetic images would require additional effort in annotation, and manual annotation can be time-consuming. Future work will consider leveraging automatic labeling algorithms, particularly those capable of handling turbid underwater conditions, to streamline the annotation process. Additionally, improving the realism of synthetic images through prompt optimization could enable LVLMs to generate more accurate and diverse underwater scenes. While we used DALL-E 3 for this work, exploring other LVLMs or generative models could offer new avenues for generating more realistic underwater images. Future research will focus on these directions, incorporating advanced generative techniques and leveraging automated annotation tools to enhance dataset quality and model performance.

\subsection{Adaptation of Florence-2 LVLM to UOD using Low Rank Adaptation (LoRA)}

\subsubsection{Introduction to Florence-2 Model}
\begin{figure}[h]
  \centering
  \includegraphics[width=0.8\linewidth]{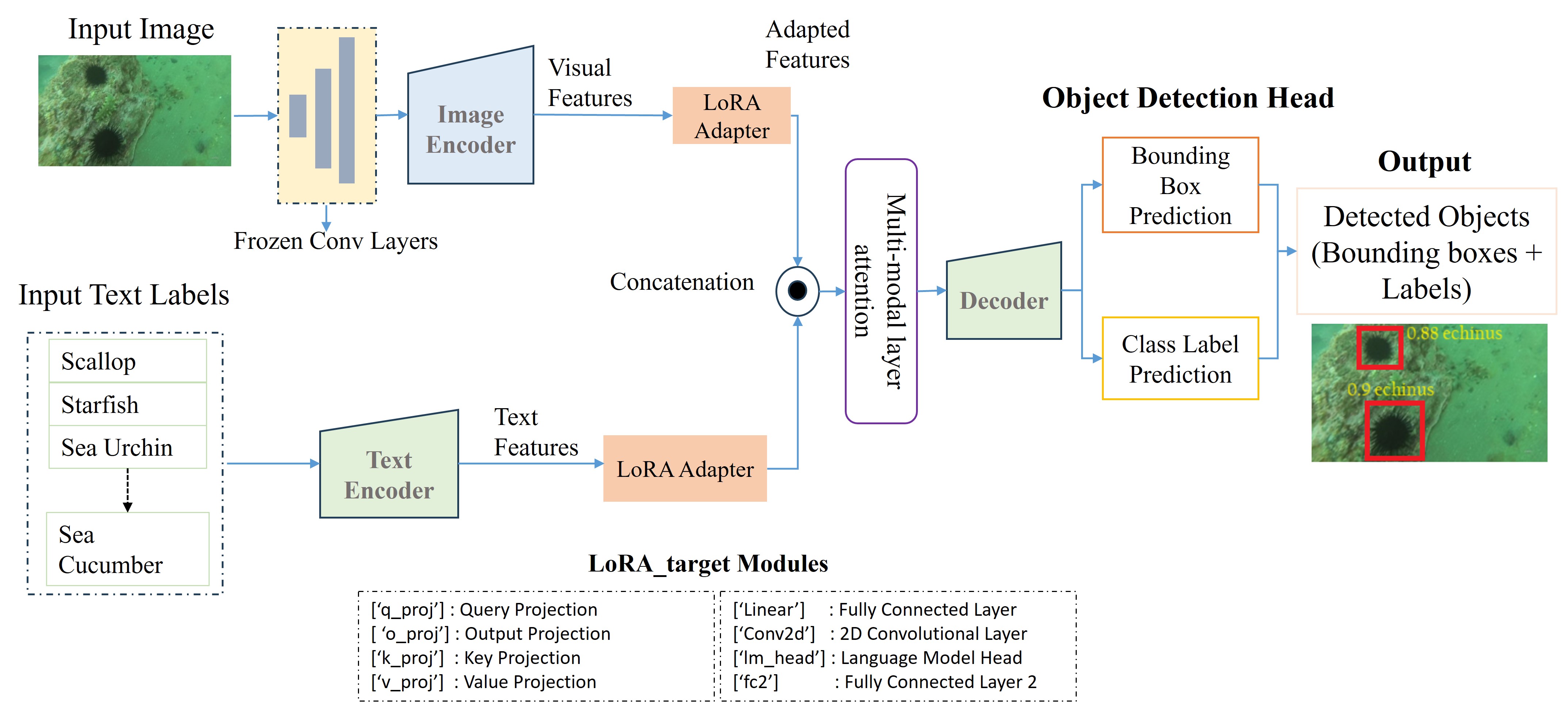}
  \caption{The Florence-2 Architecture with Highlighted Layers Where LoRA is Applied}
  \label{fig:florence-arc}
\end{figure}

Florence-2 \cite{xiao_florence-2_2023} represents a cutting-edge large vision-language model (LVLM) designed for multi-modal tasks, such as object detection, image captioning, and visual question answering. Its unified transformer-based architecture integrates visual and linguistic modalities, enabling the model to handle complex tasks with high accuracy. However, fine-tuning such a model for domain-specific tasks like UOD poses significant challenges due to its massive parameter count, which demands substantial computational resources and increases the risk of overfitting.
To address these challenges, parameter-efficient fine-tuning (PEFT) methods like Low Rank Adaptation (LoRA) \cite{hu_lora_2021} offer a viable solution. LoRA enables fine-tuning by introducing low-rank matrices into specific layers, such as attention and projection layers, of a pre-trained model. This allows only a small subset of model parameters to be adjusted, while the majority of pre-trained weights remain frozen. By focusing on critical layers, LoRA significantly reduces computational load and storage requirements compared to traditional full-model fine-tuning. This approach facilitates efficient task-specific adaptation, leveraging Florence-2’s pre-trained multi-modal representations for improved performance in UOD. LoRA’s rank-constrained matrices optimize the model for new tasks without the need to adjust all parameters, as demonstrated in Fig.~\ref{fig:florence-arc}.

\subsubsection{Experimental Setup}
The fine-tuning process utilized a pre-trained checkpoint of the Florence-2 model 
("Florence-2-base-ft")
%("microsoft/Florence-2-base-ft") 
with revision 'refs/pr/6'. The experiments were conducted on an NVIDIA A100 GPU, leveraging its computational power to fine-tune the model efficiently. The dataset used for training comprised 8,800 images, including 1,200 synthetic images generated through data augmentation and 7,600 real underwater images. The dataset covered four target classes: starfish, scallop, holothurian, and echinus.

The LoRA configuration was designed to focus on specific layers within Florence-2’s architecture, including the query (q\_proj), key (k\_proj), value (v\_proj), and output projection (o\_proj) layers, as well as fully connected layers (linear) and convolutional layers (Conv2d). This configuration allowed fine-tuning only 1,929,928 parameters, accounting for 0.7076\% of the model’s total parameters.
The training process employed an AdamW optimizer with a learning rate of 5e-6 and spanned 20 epochs. The fine-tuning process, including forward and backward passes, was completed in approximately 5 hours and 42 minutes. A structured training loop was implemented, enabling Florence-2 to adapt efficiently to the underwater domain while minimizing computational costs.

\subsubsection{Results and Observations}
\begin{figure}[h]
  \centering
  \includegraphics[width=0.8\linewidth]{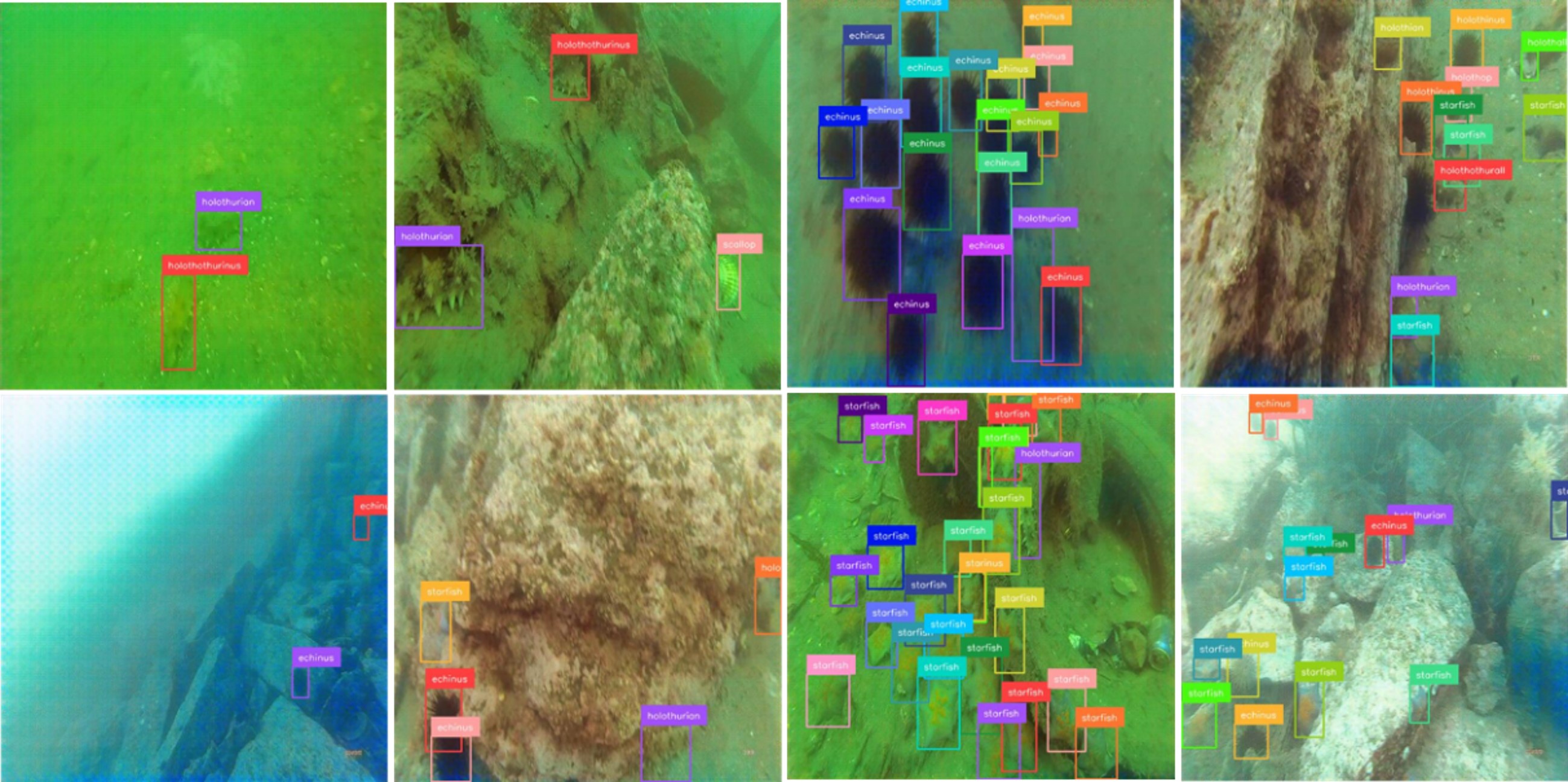}
  \caption{Fine-tuned Florence 2 Detection results}
  \label{fig:results_florence}
\end{figure}

At the end of 20 epochs, the fine-tuned Florence-2 model demonstrated strong localization capabilities, successfully detecting and localizing small underwater objects. The model excelled in drawing bounding boxes around targets, particularly for challenging classes like starfish and echinus. This highlights the potential of Florence-2 in addressing UOD challenges, especially in identifying small or occluded objects. However, some limitations were observed:

\textbf{Hallucination of Class Names:}
The model frequently generated misspelled class names, such as "echinullop," "starchin," and "hollothilicin," instead of the correct labels (echinus, starfish, holothurian, scallop). This issue severely impacted the evaluation metrics, making it impossible to compute meaningful mAP or recall values. The hallucination suggests a lack of grounding in the model’s text generation pipeline, which needs further refinement. Examples of strong localizations and hallucinations are presented in Fig.~\ref{fig:results_florence}.

\textbf{Catastrophic Forgetting:}
The fine-tuned model struggled to generalize to objects outside the training dataset. This phenomenon, known as catastrophic forgetting, is a common challenge when adapting pre-trained LVLMs to narrow domains. The model’s inability to retain its broader pre-trained knowledge limits its applicability in real-world scenarios. 

\textbf{Understanding Hallucination in Florence-2:}
Hallucination \cite{huang_survey_2025},\cite{xu_hallucination_2024} refers to a model's tendency to generate outputs that are not grounded in the input data. In the case of Florence-2, hallucination manifested as misspelled class names, reflecting an issue of ‘semantic hallucination’. This occurs when the generated output is related to the input but deviates slightly in meaning, such as predicting 'echithurian' instead of 'echinus'. 
The primary causes of hallucination in this context include limited training data, which lacks sufficient representation of rare or domain-specific terms like underwater species names; suboptimal tokenization, where rare terms are split into subwords, leading to incorrect predictions; model bias, as the Florence-2 model was pre-trained on general-purpose data with infrequent or misspelled domain-specific terms; ambiguity in multi-modal inputs, where weak correlations between visual features and class names result in incorrect outputs; and beam search bias, which favors similar but incorrect terms. These factors significantly impacted model evaluation, as hallucinated labels did not match ground truth, rendering key metrics like mAP unreliable.

\subsubsection{Strengths and Weaknesses}
The fine-tuning experiment using LoRA to adapt Florence-2 for UOD revealed both strengths and limitations. On the positive side, the parameter-efficient approach allowed fine-tuning on an A100 GPU within a manageable time frame, utilizing only 0.7076\% of the model’s total parameters, while the model demonstrated strong capabilities in localizing small objects, showcasing its potential for challenging underwater environments. However, limitations included frequent hallucinations in class names, which reduced the model’s accuracy and reliability for practical applications, and instances of catastrophic forgetting, highlighting the need for strategies to retain general knowledge while adapting to task-specific requirements.

\subsubsection{Recommendations for enhancing Florence-2 in UOD}
Mitigating hallucinations in Large Vision-Language Models (LVLMs) can be addressed through prompt tuning and engineering [110]. Carefully crafted prompts guide the model to generate accurate, contextually relevant outputs, reducing irrelevant or fabricated information. Iterative refinement of prompts further improves task alignment, enhancing reliability and generalization. Additionally, using larger models and more training data \cite{xu_hallucination_2024} can help reduce hallucinations. Larger models capture nuanced patterns better, while diverse datasets improve generalization, leading to more accurate and contextually appropriate outputs.
Instruction tuning \cite{luo_empirical_2023} reduces LLM hallucinations by fine-tuning models to follow instructions accurately. It aligns outputs with user intent, minimizes irrelevant or fabricated information, and improves coherence through task-specific training, enhancing reliability.
Improved dataset augmentation, including the use of generative models like DALL-E 3, can enhance the diversity and realism of synthetic underwater images.

\section{FUTURE RESEARCH AND CONCLUSION}

According to the discussions in this work, the future of UOD centers on overcoming persistent challenges of data scarcity, image degradation, computational constraints, and model adaptability through innovative methodologies and advanced technologies. While this study demonstrated promising yet incremental progress, such as synthetic data augmentation and LVLM fine-tuning, further research must refine and scale these approaches to achieve transformative UOD performance. Key priority directions include:

%\textbf{Future Directions}

\begin{enumerate}
\renewcommand{\labelenumi}{\theenumi)} % 在列表内部修改编号格式
\item\textbf{Efficient Fine-Tuning:}
Efficient fine-tuning techniques like adapter tuning and prompt tuning can significantly enhance the adaptability of LVLMs for UOD. Adapter tuning introduces lightweight modules to pre-trained LVLMs, enabling fine-tuning for specific underwater environments without retraining the entire model. For instance, adapters optimized for low-light, scattering, or color distortion can help adapt LVLMs to complex underwater conditions while minimizing computational overhead \cite{zhang_vision-language_2023}, \cite{xing_survey_2024}.

\item\textbf{Realistic Synthetic Data Generation:}
In the realm of UOD, generating realistic synthetic data is pivotal to mitigate challenges like data scarcity and domain adaptation. Recent advancements have highlighted the transformative potential of diffusion-based methods particularly when combined with transformer architectures in human pose estimation \cite{cui_multi-scale_2025}. More research suggest that integrating diffusion models with other generative approaches can significantly enhance the fidelity of synthetic underwater imagery \cite{yang_diffusion_2024} and  For instance, applying diffusion models within the latent space of VAEs can effectively capture intricate underwater features. Additionally, incorporating noise into the discriminator inputs of GANs can improve the realism of generated images. Furthermore, employing normalizing flows with noise injection in both forward and backward processes can refine the data distribution, leading to more authentic synthetic data. By leveraging these hybrid methodologies, future research can produce synthetic underwater datasets that closely mirror real-world conditions, thereby enhancing the performance of UOD systems.

\item\textbf{Dataset labelling:}
The Label-driven Automated Prompt Tuning (LAPT) framework offers a promising solution to reduce manual labeling efforts for underwater images. By automating prompt engineering and leveraging image synthesis and retrieval methods, LAPT eliminates the need for domain expertise and manual intervention. It autonomously generates distribution-aware prompts and training samples, improving in-distribution classification and out-of-distribution detection \cite{zhang_lapt_2024}. To accurately label small targets in underwater scenarios, advanced auto-labeling models built on frameworks like Grounding DINO and SAM \cite{mumuni_segment_2024} can be highly effective. Grounding DINO’s REC ability detects arbitrary objects using language descriptions, while SAM excels in zero-shot segmentation, saving time and improving precision. By fine-tuning these models and addressing limitations like false positives, through strategies such as size-based filtering, auto-labeling tools can minimize misses and reduce manual effort, revolutionizing underwater image analysis with robust, precise annotations.

\item\textbf{Lightweight Architectures for Real-Time Processing:}
Real-time detection is crucial for autonomous underwater vehicles (AUVs) and real-time monitoring systems\cite{guleria_systematic_2025}. 
 . LVLMs optimized for lightweight processing, such as those employing model pruning and transformer compression, can achieve this goal \cite{zhao_lightweight_2024}. 

\end{enumerate}

%\textbf{Conclusion}

In conclusion, this review has systematically examined the challenges, advancements, and emerging solutions in UOD, highlighting the transition from traditional techniques to modern deep learning approaches, including LVLMs. We categorized UOD challenges into five key areas (image quality degradation, target-related issues, data-related challenges, computational constraints, and limitations in detection methodologies), explored various image processing and object detection methodologies, and assessed the potential of LVLMs in addressing long-standing issues such as image degradation, small object detection, and dataset limitations. Furthermore, we discussed the role of synthetic data generation using DALL-E 3 and the fine-tuning of Florence-2 LVLM for UOD tasks, demonstrating both the promise and current limitations of these models.
While LVLMs offer significant potential for enhancing UOD, their real-time application and optimization for underwater environments remain underexplored. Additionally, synthetic data generation still requires refinements to ensure greater realism and applicability. Future research should focus on improving generative models for underwater-specific synthesis, optimizing LVLMs for efficient deployment, and exploring hybrid approaches that combine multiple techniques for robust detection in complex underwater settings. By addressing these gaps, UOD can be further improved to support critical applications in oceanography, marine conservation, and autonomous underwater exploration.

%%
%% The acknowledgments section

%%\section*{Acknowledgements}
%%This work was supported by National Natural Science Foundation of China (No. 61972240)

\section{Declarations}

\subsection*{Funding}
This work was supported by National Natural Science Foundation of China (No. 61972240)

\subsection*{Declaration of generative ai and ai-assisted technologies in the writing process}
During the preparation of this manuscript, the authors used ChatGPT to remove grammatical errors and refine the language. After using this tool, the authors reviewed and edited the content as needed and take full responsibility for the content of the published article.

%% If you have bib database file and want bibtex to generate the
%% bibitems, please use
%%
%%  \bibliographystyle{elsarticle-num} 
%%  \bibliography{<your bibdatabase>}

%% else use the following coding to input the bibitems directly in the
%% TeX file.

%% Refer following link for more details about bibliography and citations.
%% https://en.wikibooks.org/wiki/LaTeX/Bibliography_Management

\bibliographystyle{unsrtnat}  % Springer-like numeric style
\bibliography{Main-manuscript}

  %% The Appendices part is started with the command \appendix;
%% appendix sections are then done as normal sections
\appendix

\section{The Process of Image Enhancement}
\setlength{\abovedisplayskip}{0pt}
\setlength{\belowdisplayskip}{0pt}

\subsection{Aesthetic Enhancement}

In our case, the process of Underwater Aesthetic Enhancement involves a series of image manipulations which include applying a color tint, adding turbidity and light scattering effects, and simulating floating particles to mimic the suspended materials in water and finally a depth of field effect is created to enhance the sense of focus and immersion. These combined effects create a realistic and artistic representation of the unique visual characteristics encountered in submerged environments, making it ideal for visualizing underwater. This is mathematically expressed using Eq: A.1 - A.8.\\

\textbf{Mathematical Expression of Aesthetic Enhancement} \\
Add Color Tint: \\
The tinting process blends the original image $I_{\text{original}}$ with a uniform color overlay $T$. The blending is done using a weighted sum:

\begin{equation}
  I_{\text{tinted}} = \alpha \cdot I_{\text{original}} + (1-\alpha) \cdot T \tag{A.1}
\end{equation}
where \( I_{\text{tinted}} \) is the tinted image, \( I_{\text{original}} \) is the original image, \( T \) is a matrix of the same size as \( I \), filled with the tint color (20, 50, 100), and \( \alpha \) is the weight for the original image.
\\
Add Turbidity:\\
Turbidity is simulated by adding a random noise \( N \) to the tinted image:
\[
I_{\text{turbid}} = \beta \cdot I_{\text{tinted}} + (1-\beta) \cdot N \tag{A.2}
\]
where \( I_{\text{turbid}} \) is the turbid image, \( N \) is a random noise matrix with a normal distribution, and \( \beta = 0.9 \) is the weight for the tinted image.\\
\\
Add Light Scattering: \\
Light scattering is modeled using a vignette mask \(V\), which is created using a Gaussian kernel:
\[
V = 255 \cdot \frac{G}{\|G\|} \tag{A.3}
\]
where \(G\) is the Gaussian kernel matrix computed from the image dimensions, and \(\|G\|\) is the norm of the Gaussian kernel.

The scattered image is then computed as:
\[
I_{\text{scattered}} = \gamma \cdot I_{\text{turbid}} + (1-\gamma) \cdot V \tag{A.4}
\]
where \(I_{\text{scattered}}\) is the image with light scattering, \(V\) is the vignette mask generated by the Gaussian kernel, and \(\gamma = 0.9\) is the weight for the turbid image.\\
Add Particles:\\
Particles are simulated by adding a particle mask \(P\) to the scattered image. The particle mask is created by placing a random white pixel and applying Gaussian blur:
\[
I_{\text{particles}} = \delta \cdot I_{\text{scattered}} + (1-\delta) \cdot P \tag{A.5}
\]
where \(I_{\text{particles}}\) is the image with particles, \(P\) is the Gaussian blurred random particles, and \(\delta = 0.9\) is the weight for the scattered image.\\
Apply Depth of Field:\\
Depth of field is simulated by blending a blurred version of the image \(I_{\text{blurred}}\) with the original image \(I_{\text{particles}}\) using a circular mask \(M\):
\[
I_{\text{blurred}} = \text{GaussianBlur}(I_{\text{particles}}) \tag{A.6}
\]
\[
I_{\text{final}} = M \cdot I_{\text{particles}} + (1-M) \cdot I_{\text{blurred}} \tag{A.7}
\]
where \(I_{\text{final}}\) is the final image with depth of field and \(M\) is a binary circular mask centered on the image.\\
\textbf{Overall Process:}\\
The entire process can be expressed as a composition of functions:
\[
I_{\text{final}} = f_{\text{depth}} \circ f_{\text{particles}} \circ f_{\text{scattering}} \circ f_{\text{turbidity}} \circ f_{\text{tint}} (I) \tag{A.8}
\]
where \(f_{\text{tint}}\) is the tinting function, \(f_{\text{turbidity}}\) is the turbidity function, \(f_{\text{scattering}}\) is the light scattering function, \(f_{\text{particles}}\) is the particle addition function, and \(f_{\text{depth}}\) is the depth of field function.\\

\subsection{Color Transfer for Image Enhancement} 

The color transfer technique adjusts the color distribution of a source image to match that of a reference image. By converting both images to the LAB color space, the process computes the mean and standard deviation of their channels and uses this information to normalize and scale the source image. This results in a final image where the color characteristics of the source are closely aligned with the reference, creating a more consistent and visually appealing output. Eq: A.9 - A.18 mathematically express the implementation.\\

\textbf{Mathematical Expression of Color Transfer for Image Enhancement} \\
Convert both the source and reference images to LAB color space.\\
Let \(I_{\text{source}}\) and \(I_{\text{ref}}\) represent the source and reference images in the BGR color space. The conversion to LAB color space is denoted as:

\[
I_{\text{source}}^{LAB} = \text{BGR2LAB}(I_{\text{source}}) \tag{A.9}
\]
\[
I_{\text{ref}}^{LAB} = \text{BGR2LAB}(I_{\text{ref}}) \tag{A.10}
\]

where \(\text{BGR2LAB}(\cdot)\) is the function to convert an image from BGR to LAB color space. \\
Compute Mean and Standard Deviation.\\
For each channel \(c \in \{L, A, B\}\) of the LAB images, compute the mean \(\mu\) and standard deviation \(\sigma\):
\[
\mu_{\text{source}}^c, \sigma_{\text{source}}^c = \text{MeanStdDev}(I_{\text{source}}^{(\text{LAB}, c)}) \tag{A.11}
\]
\[
\mu_{\text{ref}}^c, \sigma_{\text{ref}}^c = \text{MeanStdDev}(I_{\text{ref}}^{(\text{LAB}, c)}) \tag{A.12}
\]
where \(\text{MeanStdDev}(\cdot)\) is the function to compute the mean and standard deviation of a channel. \\
Normalize the source image:
\[
I_{\text{source}}^{(\text{LAB}, c)} = \frac{I_{\text{source}}^{\text{LAB}} - \mu_{\text{source}}^c}{\sigma_{\text{source}}^c} \tag{A.13}
\]

Scale and shift by Reference statistics:
\[
I_{\text{transferred}}^{(\text{LAB}, c)} = I_{\text{source}}^{(\text{LAB}, c)} \cdot \sigma_{\text{ref}}^c + \mu_{\text{ref}}^c \tag{A.14}
\]

Clip the Values to Valid Range:\\
Clip the pixel values of the transferred image to the valid range \([0, 255]\):
\[
I_{\text{transferred}}^{(\text{LAB}, c)} = \text{Clip}\Bigl(I_{\text{transferred}}^{(\text{LAB}, c)}, 0, 255\Bigr) \tag{A.15}
\]

Merge Channels and Convert the Image Back to BGR Color Space:\\
\[
I_{\text{transferred}}^{\text{LAB}} = \text{Merge}\Bigl(I_{\text{transferred}}^{(\text{LAB}, L)}, I_{\text{transferred}}^{(\text{LAB}, A)}, I_{\text{transferred}}^{(\text{LAB}, B)}\Bigr) \tag{A.16}
\]
\[
I_{\text{transferred}}^{\text{BGR}} = \text{LAB2BGR}\Bigl(I_{\text{transferred}}^{\text{LAB}}\Bigr) \tag{A.17}
\]
where \(\text{Merge}(\cdot)\) is the function to merge LAB channels into a single image, and \(\text{LAB2BGR}(\cdot)\) is the function to convert an image from LAB to BGR color space.\\
\textbf{Overall Process:}\\
The entire color transfer process can be expressed as a composition of functions:
\[
I_{\text{transferred}}^{\text{BGR}} = \text{LAB2BGR} \circ \text{Merge} \circ \text{Clip} \circ \text{ScaleShift} \circ \text{Normalize} \circ \text{BGR2LAB}\Bigl(I_{\text{source}}\Bigr) \tag{A.18}
\]
where \(\text{Normalize}\) normalizes the source image to zero mean and unit variance, \(\text{ScaleShift}\) scales and shifts the normalized image using reference statistics, \(\text{Clip}\) clips pixel values to the valid range, \(\text{Merge}\) merges LAB channels, and \(\text{LAB2BGR}\) converts the image to BGR color space.

\subsection{Gaussian Blur Application}
Gaussian blur reduces image noise and detail by applying a Gaussian kernel. The kernel, based on a Gaussian function, assigns higher weights to central pixels and lower weights to surrounding ones. This process softens the image, resulting in a blurred effect, which is useful for reducing sharp edges and enhancing visual appearance. Mathematical equations Eq: A.19 - A.22 explain the implementation.\\

\textbf{Mathematical explanation of Gaussian blur} \\
Gaussian Kernel:\\
Gaussian kernel \(G\) is defined as:
\[
G(X,Y) = \frac{1}{2\pi\sigma^2} \exp\Biggl(-\frac{X^2+Y^2}{2\sigma^2}\Biggr) \tag{A.19}
\]
where \((X,Y)\) are coordinates relative to the center of the kernel, \(\sigma\) is the standard deviation of the Gaussian distribution (which controls the spread of the blur). For a kernel size of \((15,15)\), the kernel is a \(15 \times 15\) matrix where each element is computed using the above formula.\\
Convolution Operation:\\
The Gaussian blur is applied to the image \(I\) using convolution:
\[
I_{\text{blurred}}(X,Y) = \sum_{i=-k}^{k} \sum_{j=-k}^{k} I(X+i, Y+j) \cdot G(i,j) \tag{A.20}
\]
where \(I_{\text{blurred}}\) is the blurred image, \(I(X,Y)\) is the pixel value at coordinates \((X,Y)\) in the original image, \(G(i,j)\) is the value of the Gaussian kernel at offset \((i,j)\), and \(k\) is half the kernel size (e.g., \(k = 7\) for a \(15 \times 15\) kernel).\\
Normalization:\\
The Gaussian kernel is normalized such that the sum of all its elements is 1, to ensure that the overall brightness of the image is preserved during the convolution:
\[
\sum_{i=-k}^{k} \sum_{j=-k}^{k} G(i,j) = 1 \tag{A.21}
\]\\
\textbf{Overall Process:}\\
The entire Gaussian blur operation can be expressed as:
\[
I_{\text{blurred}} = I * G \tag{A.22}
\]
where \(*\) denotes the convolution operator and \(G\) is the Gaussian kernel. \\

\end{document}